\documentclass[12pt]{article}
\usepackage{placeins}
\usepackage{indentfirst} 

\usepackage{threeparttable}
\usepackage{booktabs}

\usepackage{ulem}

\usepackage{color}

\usepackage{setspace} 
\usepackage{geometry}
\usepackage{enumitem}
\usepackage{lipsum}
\usepackage{hyperref}
\usepackage{lineno}

\usepackage{cite}
\usepackage{amsmath,amssymb,amsfonts}
\usepackage{authblk}
\usepackage{algorithmic}
\usepackage{graphicx}
\usepackage{textcomp}
\usepackage{multirow}
\usepackage{subcaption}
\usepackage{xcolor}
\def\BibTeX{{\rm B\kern-.05em{\sc i\kern-.025em b}\kern-.08em
    T\kern-.1667em\lower.7ex\hbox{E}\kern-.125emX}}

\usepackage{cleveref}
\crefname{figure}{Fig.}{Figs.} 
\crefname{table}{Tab.}{Tabs.} 

\begin{document}

\title{FiVOS: A Fish Segmentation Algorithm Based on Interactive Video Object Segmentation and Filter Enhancement}
\author{
    Yuqing Duan \textsuperscript{a,b,c,d},
    Song Zhang \textsuperscript{a,b,c,d},
    Shili Zhao \textsuperscript{a,b,c,d}
    Xiaoyi Gu \textsuperscript{a,b,c,d}
    Yan Meng \textsuperscript{a,b,c,d}
    Daoliang Li \textsuperscript{a,b,c,d}
    Ran Zhao \textsuperscript{a,b,c,d,*}
}
\date{}

\maketitle

\noindent
\textsuperscript{a} National Innovation Center for Digital Fishery, China Agricultural University, Beijing 10083, China \\
\textsuperscript{b} Key Laboratory of Smart Farming Technologies for Aquatic Animal and Livestock, Ministry of Agriculture and Rural Affairs, China Agricultural University, Beijing 10083, China\\
\textsuperscript{c} Beijing Engineering and Technology Research Center for Internet of Things in Agriculture, China Agricultural University, Beijing 10083, China\\
\textsuperscript{d} College of Information and Electrical Engineering, China Agricultural University, China Agricultural University, Beijing 10083, China\\
\textsuperscript{*} Corresponding author at: P. O. Box 121, China Agricultural University, 17 Tsinghua East Road, Beijing 100083, China. 
E-mail address: ran.zhao@cau.edu.cn (R. Zhao)\\

\begin{abstract}
With the continuous expansion of aquaculture, precise and efficient monitoring of fish behavior has become increasingly critical for improving farming efficiency and reducing economic losses. In particular, with the ongoing enhancement of computational capabilities in deep learning models, vision-based fish segmentation methods are garnering growing attention. By analyzing video segmentation results, fish behavior can be effectively tracked, thereby providing reliable data support for the precise regulation of aquaculture environments.
However, existing deep learning-based video segmentation methods for aquaculture scenarios often overlook the dynamic correlations between video frames. In contrast, Interactive Video Object Segmentation (IVOS) employs an interaction-propagation scheme to achieve high-precision segmentation while minimizing user effort, thereby enhancing monitoring efficiency. Yet, IVOS applications in aquaculture remain limited due to data scarcity, and are susceptible to error accumulation and mask loss over long sequence propagation due to high intra-class similarity. In response, this paper proposes an improved interactive video object segmentation method (FiVOS) and constructs two fish-specific datasets. FiVOS utilizes a mask block filter to enable early detection and correction of erroneous propagated mask blocks, enhancing filtering accuracy through a rule-based thresholding approach. Additionally, it serializes noise filters to further eliminate erroneous mask noise, thereby improving model robustness. Experimental results demonstrate that FiVOS achieves state-of-the-art (SOTA) performance in fish video segmentation tasks, providing robust technical support for fish behavior research.

\end{abstract}


\doublespacing

\section{Introduction}

Information on fish characterization not only provides essential data support for the aquaculture sector, helping scientists and researchers better understand fish growth, health conditions, and behavioral patterns to optimize aquaculture management and resource allocation \cite{vieira2023assessment,mandal2024role}, but also serves as foundational data for studying fish adaptability in ecological environments, contributing to a deeper understanding of fish physiology and behavioral characteristics \cite{vabo2021automatic}.

However, conventional methods, such as manual observation, direct identification, and sensory evaluation, face notable limitations in data acquisition efficiency and accuracy, and may even adversely affect fish welfare and growth. Thus, there is an urgent need for advanced techniques to efficiently extract fish characterization information. The rise and development of computer vision and artificial intelligence technologies have introduced efficient and non-invasive solutions for extracting fish characterization information.
In recent years, researchers have conducted extensive exploration of image data through computer vision technologies, using techniques such as segmentation, object detection, and object tracking \cite{long2015fully,redmon2016you,nam2016learning}. These techniques have initially facilitated the development of various fish-related models, such as fish identification \cite{liu2024dp,7026059}, behavior analysis \cite{zhao2024multilevel,wang2022real}, and quantity estimation \cite{zhang2020automatic}, among others \cite{zhao2022lightweight,li2024yolo,sun2022water}.
Meanwhile, the integration of video data has further propelled research in fish characterization. Video data holds significant value not only in fish behavior analysis, health monitoring, and ecosystem observation but also in unveiling fish behavior patterns, social structures, and ecological adaptability across different environmental conditions \cite{delcourt2013video,zhao2024vision}. Through the analysis of fish video data, researchers can gain a deeper understanding of fish stress responses, feeding habits, and reproductive behaviors, establishing a critical foundation for more precise tracking and analysis of fish activity patterns.
Although fish characterization research has made substantial progress in aquaculture and ecology, significant challenges remain in efficiently and accurately extracting dynamic behavior features of fish, particularly in complex aquatic environments. Consequently, this study presents an interactive fish video object segmentation method tailored for aquaculture scenarios, designed to enhance the accuracy and efficiency of fish characterization information extraction. This method offers a novel tool for identifying and analyzing fish behavior and health status, holding significant implications for intelligent aquaculture management.

In recent years, fish behavior monitoring has gained significant attention in aquaculture and ecological conservation. Researchers have developed various segmentation, recognition, and tracking models to extract fish characterization information from video data, addressing the needs of automated monitoring and behavior analysis. These studies have laid a solid foundation for the development of algorithmic models \cite{yang2021computer}.
Rapid identification of abnormal behavior in fish is crucial for aquaculture management. Early fish behavior monitoring research was largely based on traditional computer vision techniques such as frame differencing, background subtraction, and optical flow \cite{sharif2015simple,ye2016behavioral}. However, these methods have difficulty capturing fish motion information along the depth axis, which affects the accuracy of behavior analysis. For instance, VM Papadakis et al. \cite{papadakis2012computer} employed background subtraction to detect changes in fish speed and position, analyzing escape and net-biting behaviors. Nonetheless, because it is limited to a two-dimensional plane, this method could not fully leverage the depth information of fish swimming, a crucial behavioral characteristic.
To enhance the comprehensive capture of fish swimming information, Pedersen et al. \cite{pedersen20203d} developed an RGB 3D video dataset of zebrafish and applied background subtraction and head localization to detect fish head positions, enabling three-dimensional trajectory reconstruction. However, this method is prone to target loss during detection and tracking, which decreases detection accuracy.
To tackle this challenge, Weiran Li et al. \cite{li2024tfmft} developed a multi-fish tracking model (TFMFT) based on the Transformer architecture \cite{vaswani2017attention}, which effectively overcomes tracking loss in complex backgrounds, thereby enhancing accuracy in individual trajectory monitoring and health evaluation. Although these methods have achieved satisfactory results, their ability to identify local abnormal behaviors remains limited. To address this issue, Jian Zhao et al. \cite{zhao2018modified} employed a combination of an improved Motion Influence Map and a Recurrent Neural Network (RNN) to systematically detect, locate, and identify local abnormal behaviors of fish in intensive aquaculture environments, markedly improving the accuracy of local abnormal behavior detection in dense conditions. In summary, fish behavior monitoring technology has progressed from traditional methods to extensive applications of deep learning, offering effective tools for precise monitoring in complex aquatic and dense aquaculture environments, thus promoting the realization of automation and efficiency.

Video Object Segmentation (VOS) is designed to perform high-quality foreground/background segmentation of target objects in video sequences, with wide applications in video analysis, comprehension, and editing.
Based on the degree of user intervention, VOS methods are divided into Automatic Video Object Segmentation (AVOS), Semi-automatic Video Object Segmentation (SVOS), and Interactive Video Object Segmentation (IVOS). 
IVOS facilitates the segmentation process with simple user interactions (e.g., points, scribbles, or bounding boxes), achieving efficient segmentation of target objects at a low annotation cost, which offers distinct advantages in flexibly segmenting specific objects.
In earlier studies \cite{benard2017interactive}, IVOS was implemented by combining two separate modules: an interactive image segmentation model \cite{xu2016deep,sofiiuk2020f} to generate a target mask for a single frame based on user annotations, and an SVOS model \cite{caelles2017one,oh2019video} to propagate this mask from the annotated frame to other frames. Later, SW Oh proposed a more compact solution \cite{oh2019fast}, where the model still used two separate modules for interaction and propagation, but incorporated internal connections through intermediate feature exchange and introduced external conditional links, enabling cooperation between the two modules. In IVOS-ATNet \cite{heo2020interactive}, this design was continued, with further optimization of the propagation module to separately handle tracking and propagation of local (neighboring frames) and global (distant frames) masks. However, these methods \cite{benard2017interactive,heo2020interactive} require re-running forward computations in each interaction round, leading to a gradual decline in efficiency as interaction rounds increase. To enhance efficiency, MA-Net \cite{miao2020memory} proposed a more efficient solution, centered on generating pixel embedding features through a unified encoder at the initial stage, and introducing two small branch networks for interactive segmentation and mask propagation. This model extracts pixel embeddings for all frames only in the initial round, while subsequent rounds perform forward computations solely within the two shallow branches, significantly boosting interaction efficiency. In summary, IVOS enables real-time interaction between user and task, offering high flexibility, rapid response, and low annotation costs, making it a practical solution for fish behavior detection.

\begin{figure}[ht!]
\centering
\includegraphics[width=0.9\linewidth]
{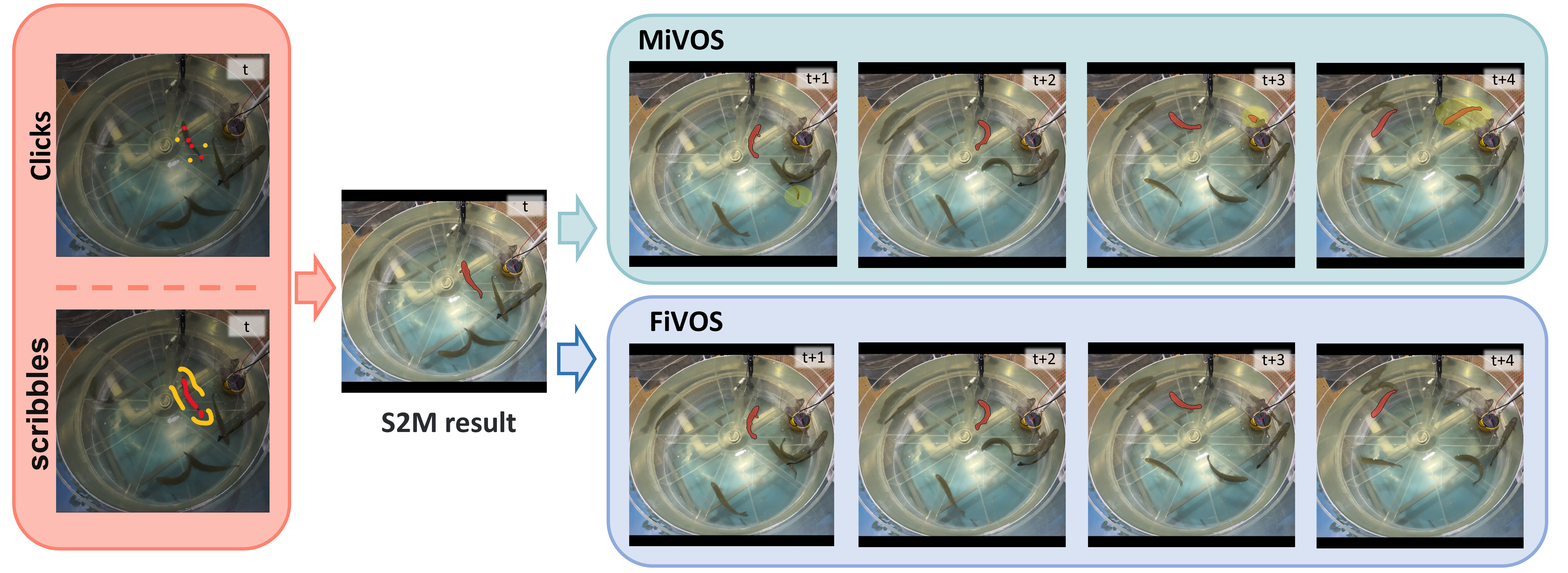}
\caption{Users annotate a single frame with simple clicks or scribbles 
(red for positive, yellow for negative), generating a high-precision mask 
via Scribble-to-Mask (S2M). The propagation module thenpropagates the mask throughout the video sequence. The two rows on the right compare the propagation results of the baseline model and FiVOS. Unlike the baseline, which suffers from irreversible error accumulation, FiVOS effectively mitigates this issue.
}
\label{fig:introdact}
\end{figure}

Current methods for fish behavior analysis do not deeply utilize dynamic correlations between video frames and lack specific object focus, making it prone to interference when capturing individual fish behavior characteristics. This limitation is particularly evident in multi-object environments, where target confusion and feature loss are common issues.
In comparison, interactive video object segmentation provides an advantage by focusing on specific objects, allowing for more targeted tracking.
Nevertheless, while current interactive video object segmentation algorithms excel on public datasets, they encounter challenges in fish farming environments, not only due to data scarcity but also from interference by complex factors such as water flow and lighting, which make high precision difficult to sustain. Even after multiple rounds of interactive correction, segmentation accuracy remains limited.
Furthermore, in fish farming, it is common to concentrate fish of the same species and growth stage within the same environment. This practice leads to high intra-class similarity between individuals, making it difficult to distinguish them during propagation and intensifying the error accumulation effect in interactive video segmentation.
As illustrated in \Cref{fig:introdact}.
Interactive video object segmentation also faces the problem of “catastrophic forgetting”, which refers to the potential loss of target masks during long-sequence propagation.

To address the aforementioned issues, this paper introduces an innovative interactive video object segmentation method (FiVOS) and develops two fish-specific datasets for training FiVOS.
The first dataset contains static images of 1,350 fish instances, offering a wealth of fish appearance features.
The second pseudo-dynamic dataset comprises 23 image sequences, with each sequence containing 10 consecutive frames to capture fish motion characteristics.
Both datasets include pixel-level mask annotations, offering precise supervisory information for model training.
In the FiVOS method, to alleviate error accumulation due to high intra-class similarity, a specialized mask block filter is designed, enabling the model to detect and filter erroneous mask blocks early in the propagation process.
To further eliminate residual erroneous noise potentially left by the mask block filter, this method incorporates a noise filter equipped with median filtering to remove erroneous mask noise, thereby enhancing the model's robustness.
Finally, a rule-based thresholding method is employed to set the optimal distance threshold, enhancing the accuracy of erroneous mask detection.
The contributions of this paper include the following:

\begin{enumerate}
    \item This paper proposes an interactive video object segmentation method for aquaculture, termed FiVOS. This method achieves state-of-the-art (SOTA) performance on fish videos, offering a powerful tool for fish behavior feature analysis.

    \item A novel filtering scheme is introduced for the detection and correction of erroneous mask blocks. This scheme comprises a serial arrangement of a mask block filter and a noise filter, which enhances the accuracy of generated masks by refining dynamically generated coarse masks during propagation. It substantially improves segmentation performance for tasks with high intra-class similarity and in challenging environments.

    \item Two fish-specific datasets were developed for FiVOS training: a static image dataset containing 1,350 fish instances, and a pseudo-dynamic dataset comprising 23 image sequences extracted from videos, with each sequence containing 10 consecutive frames to capture local movement characteristics of fish.
\end{enumerate}

\section{Materials and methods}
\subsection{Dataset acquisition}

\begin{figure}[ht!]
\centering
\includegraphics[
  width=0.8\textwidth]
{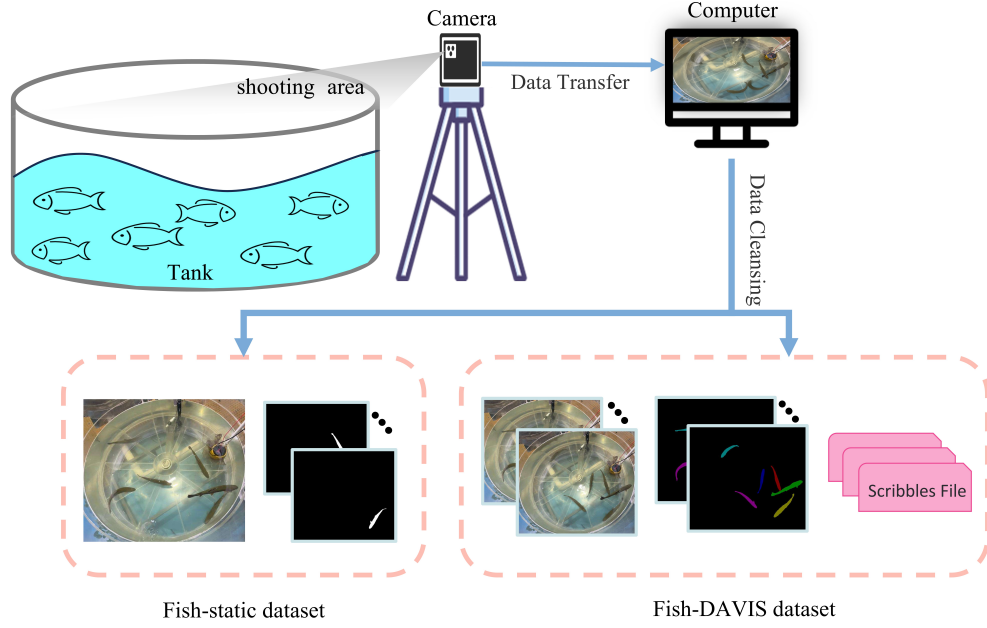}
\caption{Data Collection Schematic.
Videos of fish swimming in a tank are captured using a high-resolution camera, covering the entire water surface. The videos are then processed on a computer to generate the Fish-static and Fish-DAVIS datasets.}

\label{fig:data-env}
\end{figure}

The experimental data was collected at the National Innovation Center for Digital Fishery. 
As illustrated in \cref{fig:data-env}, a smartphone equipped with 48MP main camera, 4K video recording at 60 FPS and optical image stabilization 
was used to capture clear and stable video footage under different lighting and aquatic conditions. The tank used for the experiments is a circular acrylic tank with a diameter of 1.5 m and a depth of 0.4 m, specifically designed to provide a stable and controlled environment for the fish.

In this study, rainbow trout (Oncorhynchus mykiss) was selected as the observation species due to its status as one of the earliest domesticated economic fish, which benefits from well-established aquaculture practices and serves as a benchmark for behavioral research, and its streamlined body facilitates the precise tracking of movement patterns (e.g., swimming trajectories and postural changes), thereby enabling the construction of a high-precision dataset under controlled conditions \cite{fiordelmondo2020improvement}. During data collection, we ensured that the fish were able to move freely within the tank, allowing the data to represent their natural behavior.

To further ensure the stability of the experimental environment and the reliability of the experimental data, we monitored and recorded the key water quality parameters in the experimental setup, including water level, water temperature, pH, and dissolved oxygen concentration. These parameters play a crucial role in influencing fish behavior and health. Therefore, throughout the entire experimental process, we ensured that the water quality remained within an appropriate range.
As shown in \cref{tab:Water Quality Parameters}, the water quality parameters in the experimental environment exhibited minor fluctuations and were within the suitable range for the growth and behavioral studies of rainbow trout. The average water temperature was 21.85°C ± 1.04°C, which is close to the optimal growth range for rainbow trout (13°C to 18°C) and still within its tolerable range (0°C to 25°C). Additionally, the average dissolved oxygen concentration was 9.13 ± 1.21 mg/L, consistently exceeding the minimum dissolved oxygen requirement for rainbow trout growth (5 mg/L). The pH value averaged 7.44 ± 0.18, which is near neutral and conducive to the normal living conditions of rainbow trout.

\begin{table}[!htbp]
\centering
\caption{Experimental environment water quality parameters.} 
\label{tab:Water Quality Parameters} 
    \begin{tabular}{ll}
    \toprule 
        Parameter & Value \\ 
    \hline
        Water level (m) & 0.327 ± 0.059\\ 
        Water temperature (°C) & 21.85 ± 1.04 \\ 
        pH & 7.44 ± 0.18 \\ 
        Dissolved oxygen (mg/L) & 9.13 ± 1.21 \\ 
    \hline
    \end{tabular}
\end{table}

During filming, the camera was positioned above and to the side of the tank, ensuring coverage of the entire pool area to maximize capture of fish activities within the tank.
Upon completion of filming, the video data was transferred to a computer via a data cable for further processing.
In the data processing phase, a Python script was used to extract video frames, capturing one frame every five frames and saving them in JPG format, thereby forming a static image sequence for subsequent annotation and model training.
Then, each frame's fish contours were manually annotated at the pixel level using the AnyLabeling tool, ensuring high-quality data support for the IVOS task.

To address the issue of data scarcity, two datasets have been proposed for fish farming scenarios represented by adult rainbow trout: Fish-static and Fish-DAVIS. Their configuration details are shown in \cref{tab:dataset-set}.

\begin{table}[!htbp]
\centering
\begin{threeparttable}
\caption{Configuration details of the dataset.} 
\label{tab:dataset-set} 
    \begin{tabular}{l|ll}
        \toprule 
            Dataset Name & Configuration & value \\
        \hline
            $Fish-static$ 
            & Total Samples & 1350\\
            & Image : Mask set & 1:$n_o$ \tnote{*}  \\ 
        \hline
            $Fish-DAVIS$ 
            & Total Samples & 23\\
            & Data sample frame number & 10 \\
            & Train : Val set & 18:5 \\
        \hline
    \end{tabular}
    
\begin{tablenotes}    
\footnotesize               
\item[*] $n_o$ denotes the number of target instances in the original image
\end{tablenotes}            

\end{threeparttable}
\end{table}

The Fish-static dataset is a static image dataset. Each original image may contain multiple fish targets; in the dataset, the same original image is presented as multiple instances, each instance containing only one target mask.
We annotated a total of 225 images, encompassing 1,350 instances.

The Fish-DAVIS dataset is a standardized fish dataset specifically designed for interactive video object segmentation tasks \cite{perazzi2016benchmark}.
The dataset is composed of multiple video streams, sampled every 5 frames, with a complete data sequence constructed every 10 frames to ensure inter-frame continuity and simplify data processing.
Each set of image data is associated with a set of ground truth instance masks and three randomly selected single-frame scribble annotation files, supplying adequate scribble annotations and supervisory information for the model's propagation and fusion modules.
Furthermore, the dataset includes a JSON file for validation purposes, offering detailed attribute information of video sequences (such as frame count, number of interaction files, object count, frame dimensions, etc.) to support evaluation requirements.

\subsection{Proposed method}
\subsubsection{FiVOS Overview}
This study builds upon MiVOS \cite{cheng2021modular} and employs a "three-stage" modular structure for derivation and training. The three modules are the Interaction Module, the Propagation Module, and the Fusion Module.

\begin{figure}[ht!]
\centering
\includegraphics[
    width=\textwidth
    ]
    {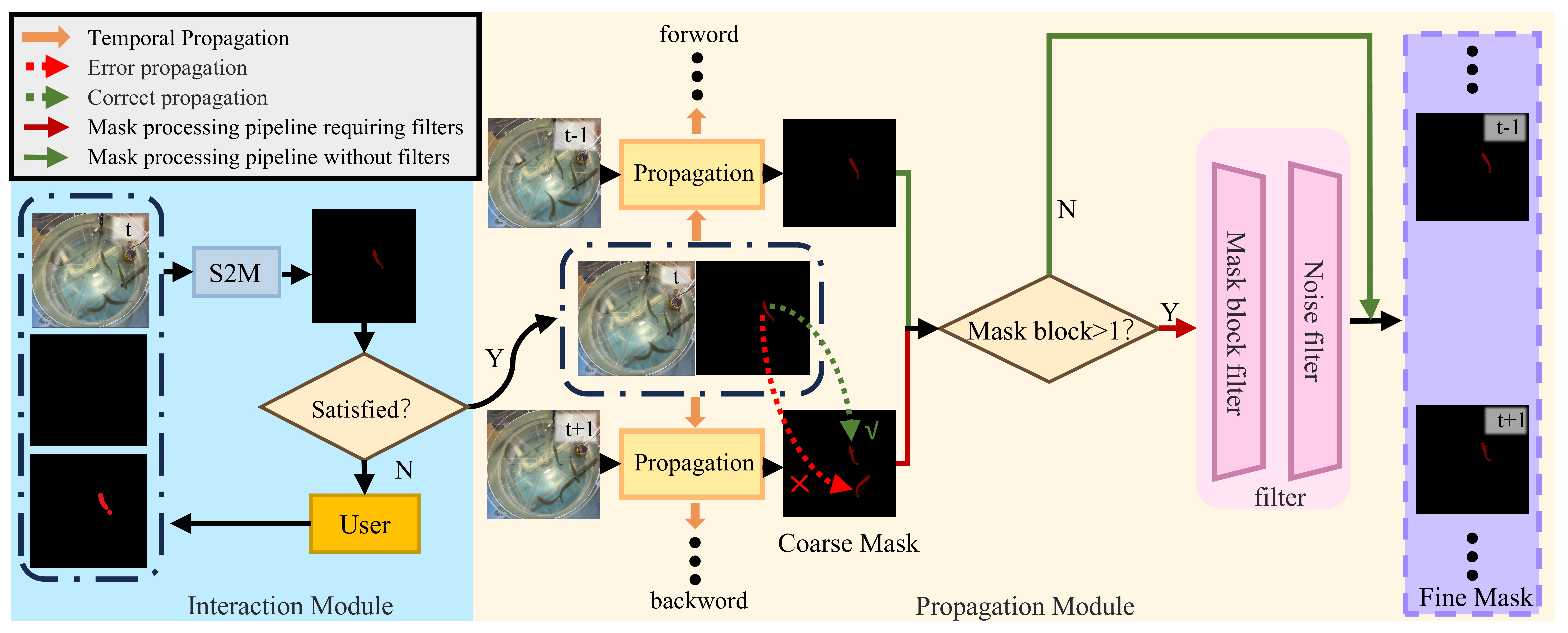}
\caption{FiVOS Framework Diagram. 
In the initial round, all masks are 
initialized to zero. The user selects frame $t$ and interactively refines the object mask using the S2M module, after which the propagation module bidirectionally propagates it throughout the video sequence to produce an initial mask.
If the initial mask contains multiple mask blocks, the filtering module removes erroneous blocks, further refining the accuracy of the predicted mask.
}
\label{fig:model_all}
\end{figure}

In the Interaction Module, the system operates within an instant feedback loop, allowing users to receive real-time feedback. This ensures satisfactory results on individual frames before moving on to the more time-consuming propagation process. In the initial round, all masks are initialized to zero. The user selects a frame $t$ and interactively refines the object mask using the Scribble-to-Mask (S2M) module until the results are satisfactory.
Next, during the mask Propagation process, the refined mask from the selected frame is bidirectionally propagated across the entire video sequence, producing an initial set of rough prediction masks. To address the issues of high intra-class similarity within datasets and exponential error accumulation, these initial prediction masks are evaluated before adjacent frame propagation. Specifically, the system assesses the number of mask blocks in the current target's prediction mask. If the predicted mask contains more than one block, the system engages a mask filter to determine whether the prediction mask includes misidentified regions. The filter removes erroneous mask blocks and generates a more accurate prediction mask. Conversely, if the predicted mask has only one block, it is assumed to be error-free and bypasses the filtering process.
Finally, the Fusion Module integrates the propagated masks with the results from the previous iteration. It captures user intent by analyzing the differences between masks selected before and after user interaction. These differences guide the fusion process to further refine the mask. Since this module is not the primary focus of our study, we will not elaborate on its details here.
The schematic of FiVOS is illustrated in \Cref{fig:model_all}.

\subsubsection{Propagation Module}

\begin{figure}[ht!]
\centering
\includegraphics[width=\textwidth, keepaspectratio]{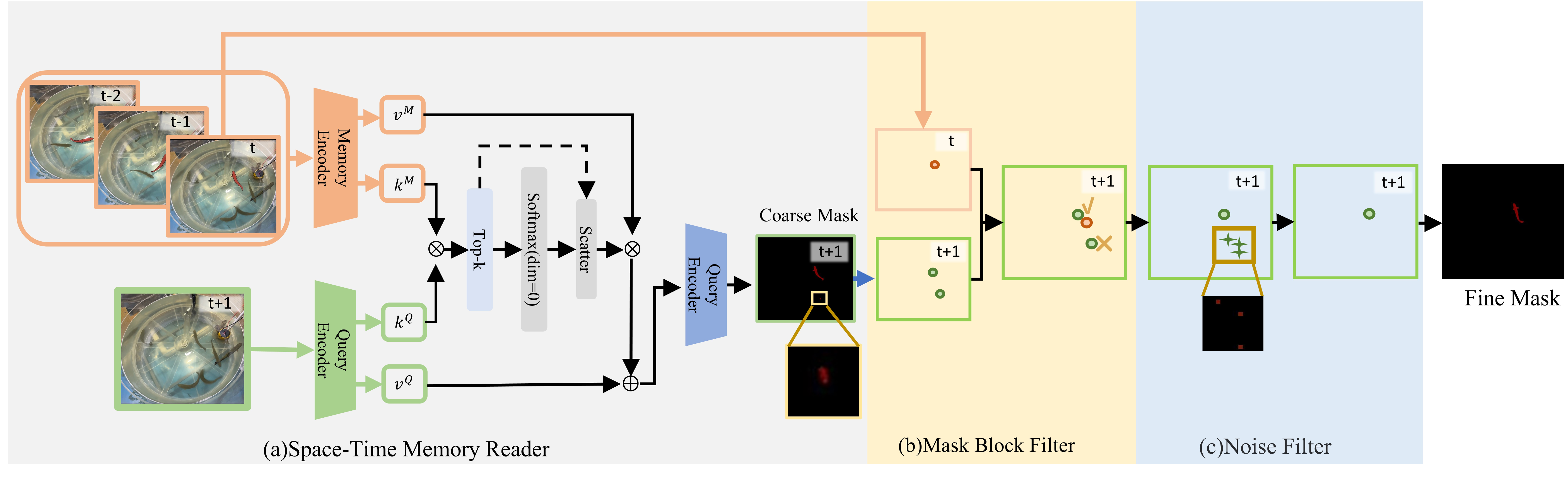}
\caption{
Propagation model architecture. (a) The Space-Time Memory Reader generates an initial prediction mask for the query frame using memory frames. (b) The Mask Block Filter identifies and removes misidentified regions from the initial prediction mask. (c) The Noise Filter eliminates residual noise to further refine the mask.}
\label{fig:model_prop}
\end{figure}

The Propagation Module is a hybrid mask prediction process, integrating both soft processing (via a space-time memory reader) and hard processing (through a mask block filter and noise filter) \cite{cheng2021modular}.
The structure of the propagation model is illustrated in \cref{fig:model_prop}.
Given one or more object masks, the propagation module tracks the objects and generates corresponding initial prediction masks for subsequent frames.
Using STM with a Top-k operation \cite{lu2020video,oh2019video,luiten2018premvos}, past frames containing object masks are treated as memory frames, which are employed to predict the object mask in the current (query) frame via an attention-based memory reading operation.
It is worth noting that to address STM error accumulation caused by high intra-class similarity within the fish dataset, we propose a novel mask filter. This simple yet effective method filters out unreasonable mask blocks within the query mask via a mask block filter, ensuring the quality of memory masks in the subsequent propagation stage and alleviating target confusion and exponential error accumulation resulting from high intra-class similarity.
Even with minor recognition errors during propagation, the mask filter suppresses distant erroneous mask blocks, enabling the network to correct these errors in subsequent propagation steps and preventing the further spread of erroneous information.
Following the mask block filter, there may still be error specks in the mask, as individual error specks may not qualify as complete mask blocks. Thus, we introduced a noise filter to eliminate these potential error specks, thereby further enhancing the quality of the query mask.
Because this mask filter performs hard-processing on mask blocks, it can be flexibly integrated into various SVOS models.

\subsubsection{Space-time memory reader} 
The space-time memory reader \cite{oh2019video} module receives the initial mask input of the target object, tracks the specified object, and generates corresponding prediction masks in subsequent frames.
As illustrated in \cref{fig:model_prop} (a), this module has two input branches: one for memory frames (past frames) with object masks and one for query frames (current frames) without object masks.
Memory and query frames are passed through dedicated encoders to generate key and value mappings. The memory encoder receives both the image and the object mask, while the query encoder receives only the image.
For $T$ memory frames, key-value features are computed and then multiplied via dot product to generate a similarity matrix $F$, indicating the similarity between query positions and memory positions.
To enhance efficiency and minimize memory usage, a Top-$k$ filter retains only the top $k$ entries with the highest similarity values.
The filtered similarity matrix $W$ is multiplied by $v^M$ to produce feature $m$, which is then concatenated with $v^Q$ and sent to the decoder to generate the initial object mask.
The Scatter operation distributes data to specified index positions, forming a new similarity feature matrix $W$, thereby further improving computational efficiency and feature alignment.

\subsubsection{Mask block filter} 

\begin{figure}[ht!]
\centering
\includegraphics[width=\linewidth]{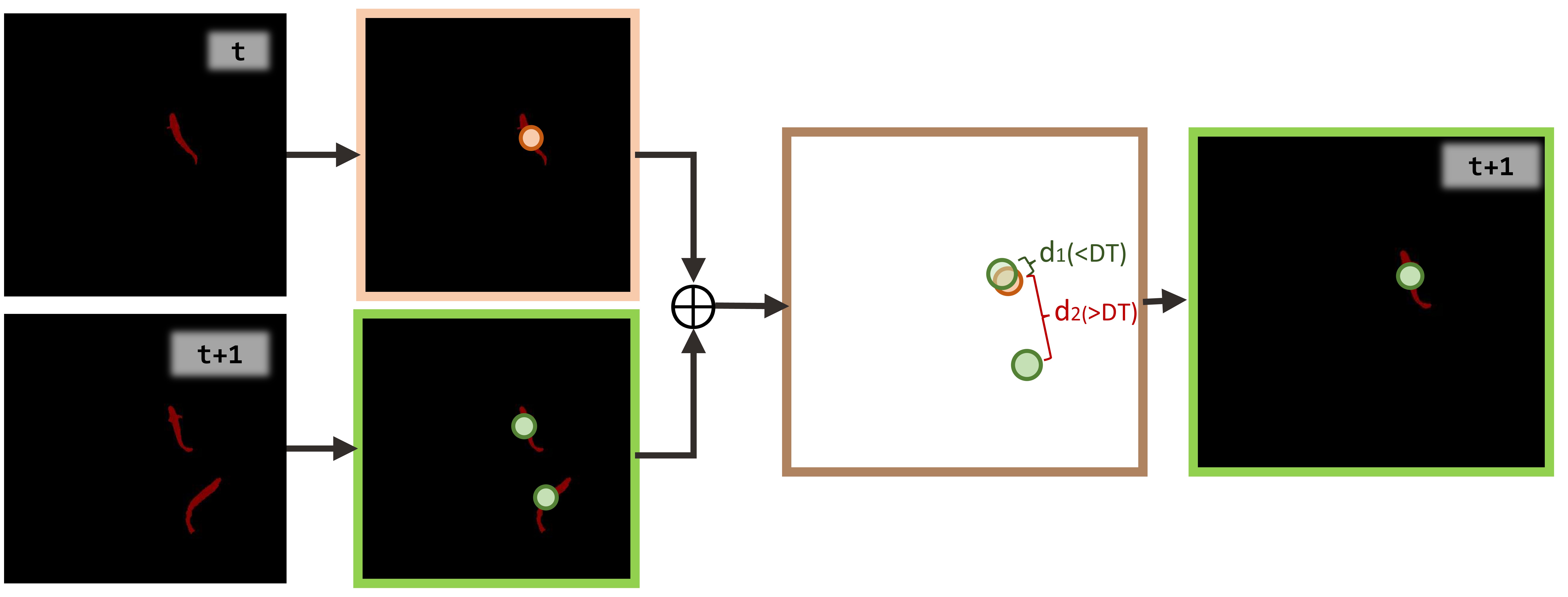}
\caption{
Mask block filter.
The centroid of each mask block is calculated, and the distance $d_n$ (with $n$ as the mask block index) between the centroids of query and memory frame mask blocks is measured. Mask blocks with $d_n>DT$ are marked as ecognition errors and set to background; others are preserved.}
\label{fig:filter1}
\end{figure}

Given the continuity of information in video streams, it is improbable for the centroid distances of mask blocks in adjacent frames to show sudden increases.
Based on this characteristic, we manually set a task-specific threshold using prior knowledge (For specific details, see Section 2.2.5). This threshold can be estimated as a reference value from the existing mask information in the dataset.
This approach helps alleviate target confusion and error diffusion issues caused by high intra-class similarity.
Even with minor recognition errors during propagation, the mask filter suppresses distant erroneous mask blocks, allowing the network to correct errors in subsequent propagation stages and preventing the further spread of errors.
As illustrated in \cref{fig:model_prop} (b), the mask block filter operates by performing a many-to-many shortest distance matching between the initial object mask generated by STM and the mask blocks in adjacent frames. A minimum distance threshold is set to filter out incorrect mask blocks, ensuring the quality of memory masks in the subsequent propagation phase.
To enhance object tracking accuracy between video frames, this study calculates the maximum centroid distance within each frame as the primary reference indicator for centroid movement.
Based on this maximum distance, an empirical threshold is set to filter out frames with abnormal movement or incorrect detections.
This centroid-based maximum distance thresholding method can be regarded as a heuristic rule to promote tracking stability and accuracy in practical video processing.
As depicted in \cref{fig:filter1}, green dots indicate the centroids of predicted mask blocks in the current frame, while red dots indicate the centroids of mask blocks in adjacent memory frames.
In this example, the STM module generates two mask blocks for the current query frame, posing a risk of target recognition error.
The mask filter calculates the L2 norm (Euclidean distance) between the centroids of the predicted mask blocks and the centroids of mask blocks in adjacent memory frames, as follows:

\begin{equation}
d_{ij}=\sqrt{(x_i-x_j^{'})^2+(y_i-y_j^{'})^2}
\end{equation}

In this context, $d_{ij}$ is the distance between the $i$-th query frame mask block and the $j$-th adjacent memory frame mask block, where $(x_i, y_i)$ denotes the centroid coordinates of the $i$-th query frame mask block, and $(x_j^{'}, y_j^{'})$ represents the centroid coordinates of the $j$-th adjacent memory frame mask block.
Here, $i$ represents an integer in the range $[2, +\infty)$. Notably, when $i=1$, the mask is assumed to be free of recognition errors, thus bypassing the mask block filter. $j$ is an integer in the range $[1, +\infty)$.

\subsubsection{Rule-Based Thresholding Method}

To enhance the accuracy of object tracking across video frames, this study proposes a threshold-setting method based on the centroid distance of object masks between adjacent frames.
This method can be seen as a heuristic rule. Specifically, we define $DT$ as the maximum movement distance of object mask centroids between adjacent frames to assess centroid displacement, using it as a filter for erroneous detection to maintain tracking stability and accuracy.

The centroid represents the geometric center of the image, and we use it to denote the positional information of the target object. When calculating the centroid, we extract each target as a binary image channel and then calculate the centroid by computing the first-order moment of the pixels in each channel. The calculation formula is as follows:

\begin{equation}
C_i(x,y) = (\frac{M_{10}}{M_{00}},  \frac{M_{01}}{M_{00}})
\end{equation}

Here, \(M_{00}\) denotes the zeroth-order moment of the mask, indicating the total number of pixels within the mask area, while \(M_{10}\) and \(M_{01}\) are the first-order moments in the horizontal and vertical directions, respectively.
More specifically, the definitions of the zeroth-order and first-order moments are as follows:

\begin{equation}
M_{00} = \sum_{j} \sum_{k}I(j,k)
\end{equation}

\begin{equation}
M_{10} = \sum_{j} \sum_{k}j \cdot  I(j,k)
\end{equation}

\begin{equation}
M_{01} = \sum_{j} \sum_{k}k \cdot  I(j,k)
\end{equation}

Here, \(I(j,k)\) denotes the value of the pixel at \((j,k)\) while \(M_{10}\) and \(M_{01}\) represent the horizontal and vertical first-order moments, respectively. Subsequently, the centroid distance \(d_{t,t+1}\) between adjacent frames \(t\) and \(t+1\) can be defined as:

\begin{equation}
d_{t,t+1}=\sqrt{(C^x_t - C^x_{t+1} )^2 + (C^y_t - C^y_{t+1} )^2}
\end{equation}

Based on this definition, we define the maximum distance threshold $DT$ as:

\begin{equation}
DT=max(d_{t,t+1})
\end{equation}

In this experiment, the optimal filtering distance threshold for this task was determined to be \(DT=25.7\), based on the mask information in our Fish-DAVIS dataset.

\subsubsection{Noise filter}

As illustrated in \cref{fig:model_prop} (c), after the mask block filter has processed the mask, small error specks of noise may still persist. These noise specks are generally too small to be identified as independent mask blocks, yet they can negatively impact the accuracy of subsequent processing stages.
To resolve this issue, we developed a noise filter specifically designed to eliminate these potential noise specks, further improving the quality of the query mask.
This noise filter employs a commonly used median filtering technique, which effectively removes isolated noise points in the image by replacing the target pixel's value with the median of the surrounding pixels.

Median filtering is a nonlinear filtering method based on sorting, which effectively suppresses isolated noise points by replacing the target pixel value with the median of all pixel values within a local window. 
Let \(I(x,y)\) represent the pixel value of the image. In median filtering, for each pixel position \((x,y)\), a local window \(W(x,y)\) (e.g., of size \(3\times 3\) or  \(5\times 5\)) is chosen to encompass the pixel values around it.
The mathematical expression of this filter is:

\begin{equation}
I'(x,y) = median(W(x,y))
\end{equation}

Here, \(I'(x,y)\) represents the filtered pixel value, and \(median(W(x,y))\) denotes the median of the pixel values within the window \(W(x,y)\).

In the process of median filtering, the local window size \(W(x,y)\) is the only adjustable and critical parameter. The window size directly determines the strength and precision of the filtering: smaller windows can preserve more detailed information but may be insufficient to eliminate larger noise artifacts, while larger windows can more effectively suppress noise but may introduce over-smoothing effects, leading to the loss of image details. To investigate the impact of the local window size on noise filtering performance and model accuracy, we conducted experiments using different window sizes and quantitatively analyzed the model performance. The experimental results are summarized in \cref{tab:MedianFilterResults}.

\begin{table}[!htbp]
\centering
\caption{Performance comparison of different median filter window sizes.} 
\label{tab:MedianFilterResults} 
    \begin{tabular}{lll}
    \toprule 
       Window size & AUC-J\&F & J\&F \\ 
    \hline
        1 & 91.72 & 91.66\\ 
        3 & \textbf{93.84} & \textbf{93.89} \\ 
        5 & 92.54 & 92.49 \\
        11 &  91.32 & 91.25\\ 
        21 & 90.91 & 90.83 \\ 
    \hline
    \end{tabular}
\end{table}

When the window size is \(3\times 3\), the model achieves optimal performance. This indicates that, at this window size, noise particles are effectively suppressed while preserving important details.
A smaller window size (e.g., \(1\times 1\)) fails to adequately remove noise particles, thereby affecting the accuracy of subsequent processing stages. Conversely, an excessively large window size (e.g., \(21\times 21\)) results in over-smoothing, leading to a loss of critical details and a reduction in model performance.
Therefore, we selected a window size of \(3\times 3\) as the default parameter for the median filter. This choice strikes an effective balance between noise suppression and detail preservation.

The median filter’s advantage is its ability to suppress small-area noise while retaining edge details in the image, making it particularly effective for addressing salt-and-pepper noise.
This design enables our noise filter to enhance the clarity and reliability of the mask, ensuring that small noise specks do not impact the system’s overall performance and accuracy in subsequent processing stages.
This combined approach of using both the mask block filter and the noise filter guarantees stability and accuracy when handling complex video stream data.

\section{Results and Analysis}

\subsection{Experimental confguration}

The experimental process is divided into three stages, each corresponding to the derivation and training of a specific module. The overall training workflow is as follows:

\begin{enumerate}
    \item The first part is the Scribble To Mask module, which we train independently on the static image dataset Fish-static, with pre-trained weights of deeplabv3plus\_resnet50 loaded to facilitate rapid model fitting.

    \item The second part is the Mask Propagation module, with training divided into two steps. First, the model is trained on the Fish-static dataset, and the resulting weights are saved. Training then continues on the Fish-DAVIS dataset using these saved weights. In each training iteration, three frames are randomly selected from the video sequence, with the maximum distance between frames gradually increasing from 5 to 10 (curriculum learning) and annealing back to 5 at the end of training.

    \item The Fusion module uses Fusion data as input, which is generated from the output of the trained Mask Propagation module. The Fish-DAVIS video sequence dataset is used for loss calculation during training.
\end{enumerate}

The weights of the aforementioned three modules are ultimately utilized during the validation process.
All experiments were conducted on an Ubuntu server. The hardware and software configurations are shown in \cref{tab:system configuration}. Model training parameter details are presented in \cref{tab:Model configuration}. 

\begin{table}[!htbp]
\centering
\caption{Experimental system and hardware configuration.} 
\label{tab:system configuration} 
    \begin{tabular}{ll}
    \toprule 
        Parameter/Configuration & Result value \\ 
    \hline
        Operating System & Ubantu 20.04.2 \\ 
        CPU & Core i5-13600KF \\ 
        GPU & NVIDIA RTX 4070 \\ 
        Video memory & 16GB \\ 
        CUDA & 11.8 \\ 
        PyTorch & 2.2.2+cu118 \\ \hline
    \end{tabular}
\end{table}

\begin{table}[!htbp]
\centering
\caption{Model parameters configuration.} 
\label{tab:Model configuration} 
    \begin{tabular}{l|ll}
        \toprule 
            Model & Parameter/Configuration & Result value \\ 
        \hline
            Scribble To Mask
            & Epoch & 80000 \\
            & Batch size & 2 \\
            & The initial learning rate & 1e-4 \\
            & Gamma & 0.1 \\
            & Dataset & Fish-static \\
        \hline
            Mask Propagation-step1
            & Epoch & 30000 \\
            & Batch size & 4 \\
            & The initial learning rate & 1e-5 \\
            & Learning Rate Step & 25000 \\
            & Gamma & 0.1 \\
            & Dataset & Fish-static \\
        \hline
            Mask Propagation-step2
            & Epoch & 50000 \\
            & Batch size & 4 \\
            & The initial learning rate & 1e-5 \\
            & Learning Rate Step & 45000 \\
            & Gamma & 0.1 \\
            & Dataset & Fish-DAVIS \\
        \hline
            Difference-Aware Fusion
            & Epoch & 30000 \\
            & Batch size & 8 \\
            & The initial learning rate & 1e-4 \\
            & Learning Rate Step & 20000 \\
            & Gamma & 0.1 \\
            & Train Dataset  & Fusion data \\
            & Compute Loss Dataset & Fish-DAVIS \\
        \hline
    \end{tabular}
\end{table}

\clearpage

\subsection{Evaluation metrics}

In interactive video object segmentation tasks, evaluating model performance is essential, as segmentation accuracy and interaction efficiency directly impact user experience.
This study uses the $J\&F$ and $AUC-J\&F$ metrics to evaluate segmentation accuracy, edge consistency, and overall performance across different interaction counts.
These evaluation metrics are widely accepted and interpretable in the field of video segmentation.

Specifically, the $J\&F$ metric combines the Jaccard Index ($J$) and edge accuracy ($F$) to provide an overall assessment of the segmentation region’s accuracy and edge consistency.
$J$ assesses the accuracy of the segmented region by computing the Intersection over Union (IoU) between the segmentation result and the ground truth mask, while $F$ quantifies edge consistency by evaluating the alignment between the predicted and actual edges.
A high $J\&F$ value represents the model's combined performance in segmentation accuracy and edge quality.
Its calculation formula is as follows:

\begin{equation}
J = \frac{M \cap \hat{M} }{M \cup  \hat{M}}
\end{equation}

\begin{equation}
F = \frac{2P_cR_c}{P_c+  R_c}
\end{equation}

\begin{equation}
J\&F = \frac{J+F}{2}
\end{equation}

Here, \(\hat{M}\) denotes the segmentation result predicted by the model, and \(M\) is the ground truth mask.
A $J$ value closer to 1 indicates a higher degree of overlap between the segmentation result and the ground truth, reflecting the model’s effectiveness in identifying the overall contours of the target. \(P_c\) denotes the proportion of predicted edges that match the ground truth edges, while \(R_c\) denotes the proportion of ground truth edges that are correctly identified as edges.
Edge consistency is particularly crucial for segmentation tasks with high detail requirements.
An $F$ value closer to 1 implies higher edge consistency, demonstrating the model’s capacity to capture fine edge details.
A higher $J\&F$ value signifies that the model not only achieves high accuracy in the segmentation region but also maintains strong boundary consistency, allowing for more precise and nuanced evaluation of segmentation quality.

On the other hand, the $AUC-J\&F$ metric measures the model’s overall performance across different interaction steps, reflecting the area under the curve of segmentation performance relative to the number of interactions.
A higher $AUC-J\&F$ value indicates that the model can achieve stable segmentation with fewer interactions, reflecting its efficiency and stability in interactive segmentation tasks, effectively enhancing user experience.
Its formula is given as:

\begin{equation}
AUC-J\&F=\int_{t_0}^{t_n} P(t)dt
\end{equation}

Here, \(t_0\) denotes the starting interaction time point, \(t_n\) represents the ending interaction time point, and \(P(t)\) indicates the $J\&F$ value at time \(t\) or at interaction step \(t\).
The integration result, which gives the area under the curve, reflects the model's overall performance at various interaction times.

\subsection{Comparison with other advanced tracking algorithms}

To evaluate our model’s performance, the bot (i.e., the official automatic scribble API) initially provides a scribble on a selected frame and waits for the algorithm’s output.
The bot then applies corrective scribbling to the worst-performing frame among all candidate frames generated by the algorithm.
This loop can be repeated a maximum of 8 times.
Notably, since we use the private interactive dataset Fish-DAVIS, this behavior is implemented by modifying the official API’s davis.json file.
In the quantitative analysis, we applied this automated process to produce experimental results using predefined metrics. In the qualitative analysis, we conducted real-time human interaction on a 20-second long video (1200 frames) within the PyQt interface to assess user experience and visual accuracy in practical application.

The performance of our method on the Fish-DAVIS interactive validation set is presented in \cref{tab:compare-res-s}, alongside a comparison with results from current state-of-the-art methods.
All models were trained and validated on the same GPU (NVIDIA RTX 4070) to ensure fairness and consistency in the results.
The experimental results show that our method surpasses all competitors across various metrics.
However, due to the short sequence length of only 10 frames in the Fish-DAVIS interactive validation set, neither the MiVOS nor the STCN exhibited any erroneous mask blocks during this brief sequence test. As a result, no significant improvement was observed when compared to the FiVOS.
To provide a more comprehensive evaluation of model performance, we merged the Fish-DAVIS dataset into a long sequence of 80 frames for testing.
As shown in \cref{tab:compare-res-l}, in this long-sequence test, our method significantly outperformed other comparison methods, exhibiting enhanced robustness and superior segmentation accuracy.

\begin{table}[!htbp]
\caption{\label{tab:compare-res-s}Performance comparison on the Fish-DAVIS validation set (10-frame short video).} 
\centering
\begin{tabular}{ccc} 
\toprule 
 & AUC-J \&F & J \&F \\
\midrule 
MANet \cite{miao2020memory} & 67.84 & 67.90 \\
RGMP\cite{oh2018fast} & - & 69.31 \\
XMen\cite{cheng2022xmem} & - & 89.14 \\
STCN \cite{cheng2021rethinking} & 93.45 & 93.48  \\
MiVOS & 93.20 & 93.24 \\
FIVOS & \textbf{93.80} & \textbf{93.85} \\
\bottomrule 
\end{tabular}
\end{table}

\begin{table}[!htbp]
\caption{\label{tab:compare-res-l}Performance comparison on the Fish-DAVIS validation set (80-frame long video).} 
\centering
\begin{tabular}{ccc} 
\toprule 
 & AUC-J \&F & J \&F \\
\midrule 
MANet \cite{miao2020memory} & 78.63 & 79.18 \\
RGMP\cite{oh2018fast} & - & 21.19 \\
XMen\cite{cheng2022xmem} & - & 29.94 \\
STCN \cite{cheng2021rethinking} & 74.65 & 74.65  \\
MiVOS & 65.50 & 65.40 \\
FIVOS & \textbf{92.50} & \textbf{92.40} \\
\bottomrule 
\end{tabular}
\end{table}

\begin{figure}[ht!]
\centering
\includegraphics[
  width=0.8\textwidth]
{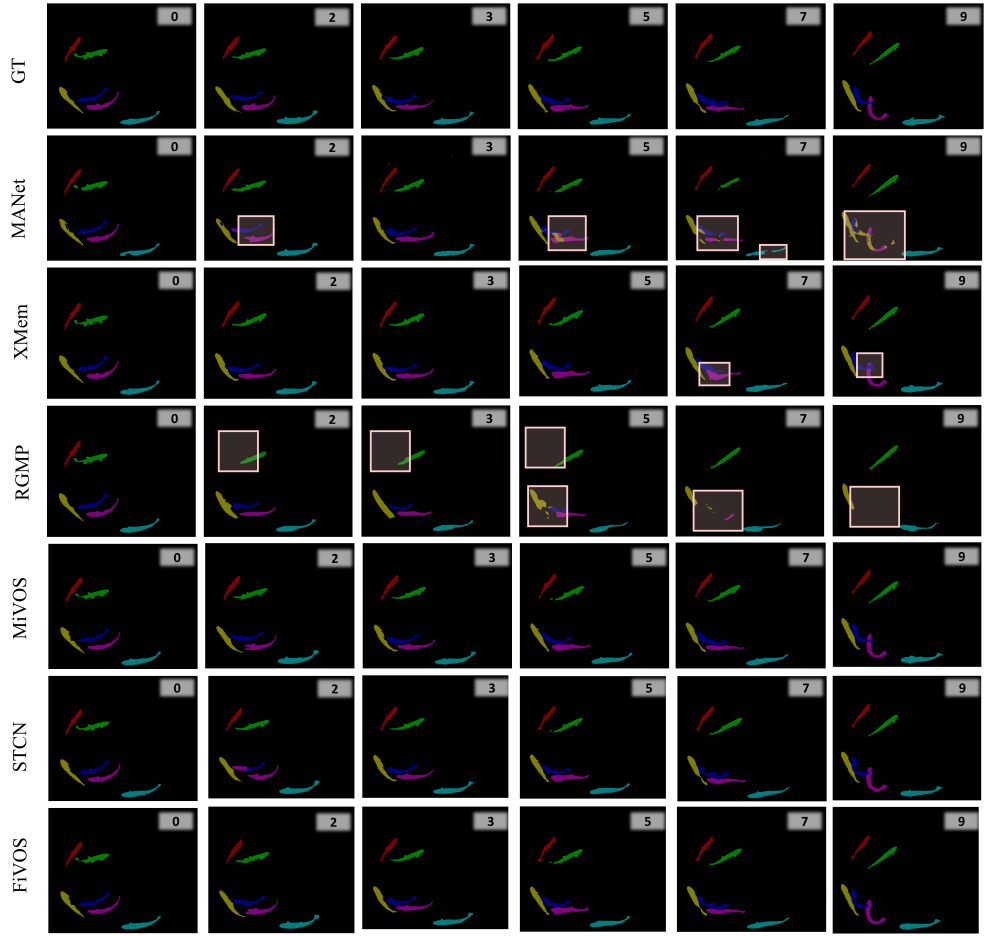}
\caption{Multi-object video segmentation example on the Fish-DAVIS validation set (10-frame short video). For a short video sequence, 
only the MANet method exhibited target confusion, while other methods produce comparable results with minor differences.}
\label{fig:compare-mask}
\end{figure}

\begin{figure}[ht!]
\centering
\includegraphics[
  width=0.8\textwidth]
{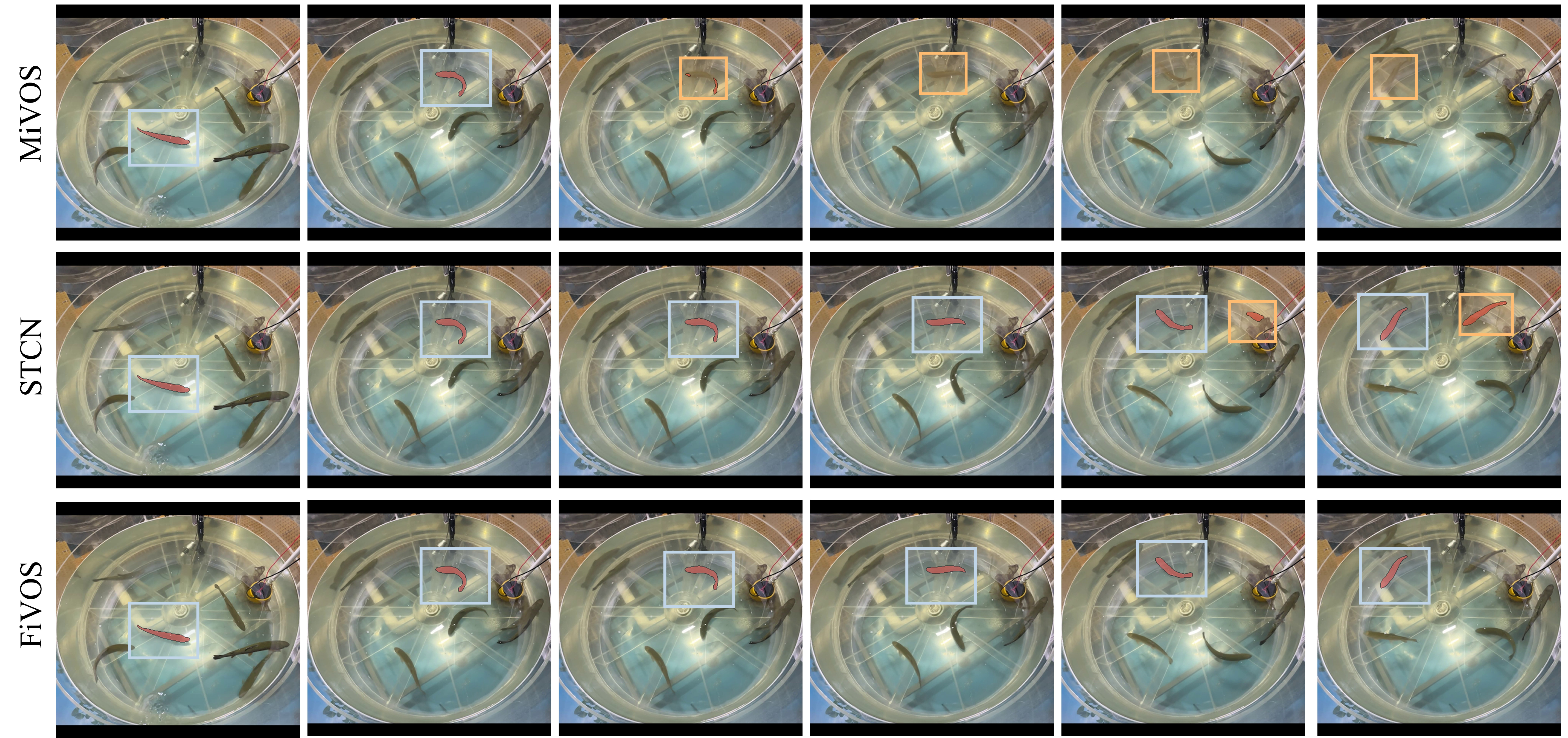}
\caption{Single-object interaction performance on a 20-second long video sequence. MiVOS and STCN experienced target loss and recognition errors at frames 99 and 117, respectively. 
In contrast, our method consistently delivers accurate segmentation results throughout the sequence.
}
\label{fig:compare}
\end{figure}

The qualitative results on the short sequences of the Fish-DAVIS dataset are shown in \cref{fig:compare-mask}. Except for MANet, XMem, and RGMP, the segmentation performance of the other three methods is comparable on short sequences (within 10 frames).
However, in the qualitative results for the long sequences (see \cref{fig:compare}), MiVOS exhibited target loss at frame 99, whereas STCN encountered a target recognition error at frame 117.
In contrast, our FiVOS algorithm maintained stable segmentation results throughout the entire process, demonstrating a notable advantage on long sequences.

\clearpage

\subsection{Ablation study}

To validate the effectiveness of each component in the proposed FiVOS, we performed ablation studies on the Fish-DAVIS dataset using both short (10-frame) and long (80-frame) video sequences.

\subsubsection{Mask block filter}

As shown in \cref{tab:self-res-s}, adding the mask block filter resulted in a slight decrease in validation accuracy on the Fish-DAVIS short video (10 frames) compared to the baseline, with a difference of 0.371\%, though there are specific reasons for this outcome.
Since recognition errors generally occur in longer videos (over 70 frames) during validation, this private dataset with only 10 frames did not reach the critical threshold for error accumulation.
In this scenario, error accumulation during validation was not severe enough to generate erroneous mask blocks, preventing the mask block filter from effectively performing its error-correction function.
However, in the Fish-DAVIS long video (80 frames), validation accuracy improved significantly after adding the mask block filter.
As shown in \cref{tab:self-res-l}, in long video validation, the method with the mask block filter improved validation accuracy by 4.6\% over the baseline model.

\begin{table}[!htbp]
\centering
\caption{\label{tab:self-res-s}Ablation study on the Fish-DAVIS short video.} 
\begin{threeparttable}
\begin{tabular}{lll} 
\toprule 
 & AUC-J \&F & J \&F \\
\midrule
Baseline & 93.198$_{\text{-}}$ & 93.242$_{\text{-}}$ \\
(+) mask block filter & 92.881$^{\downarrow \ 0.317}$ & 92.930$^{\downarrow \ 0.312}$  \\
(+) mask block filter + noise filter & 93.804$^{\uparrow \ 0.612}$ & 93.852$^{\uparrow \ 0.61}$ \\
(+) mask block filter + noise filter\tnote{+} & \textbf{93.844}$^{\uparrow \ 0.646}$ & \textbf{93.894}$^{\uparrow \ 0.652}$ \\
\bottomrule 
\end{tabular}
\begin{tablenotes}    
        \footnotesize               
        \item[+]Setting thresholds using a rule-based thresholding approach         
      \end{tablenotes}            
\end{threeparttable}
\end{table}

\begin{table}[!htbp]
\centering
\caption{\label{tab:self-res-l}Ablation study on the Fish-DAVIS long video.} 
\begin{threeparttable}
\begin{tabular}{lll} 
\toprule 
 & AUC-J \&F & J \&F \\
\midrule
Baseline & 65.5$_{\text{-}}$ & 65.4$_{\text{-}}$ \\
(+) mask block filter & 70.1$^{\uparrow 4.6}$ & 69.5$^{\uparrow 4.1}$  \\
(+) mask block filter + noise filter & 83.1$^{\uparrow 17.6}$ & 82.9$^{\uparrow 17.5}$ \\
(+) mask block filter + noise filter\tnote{+} & \textbf{92.5}$^{\uparrow 27.0}$ & \textbf{92.4}$^{\uparrow 27.0}$ \\
\bottomrule 
\end{tabular}
\begin{tablenotes}    
        \footnotesize               
        \item[+]Setting thresholds using a rule-based thresholding approach         
      \end{tablenotes}            
\end{threeparttable}
\end{table}

\begin{figure}[ht!]
\centering
\includegraphics[
  width=0.8\textwidth]
{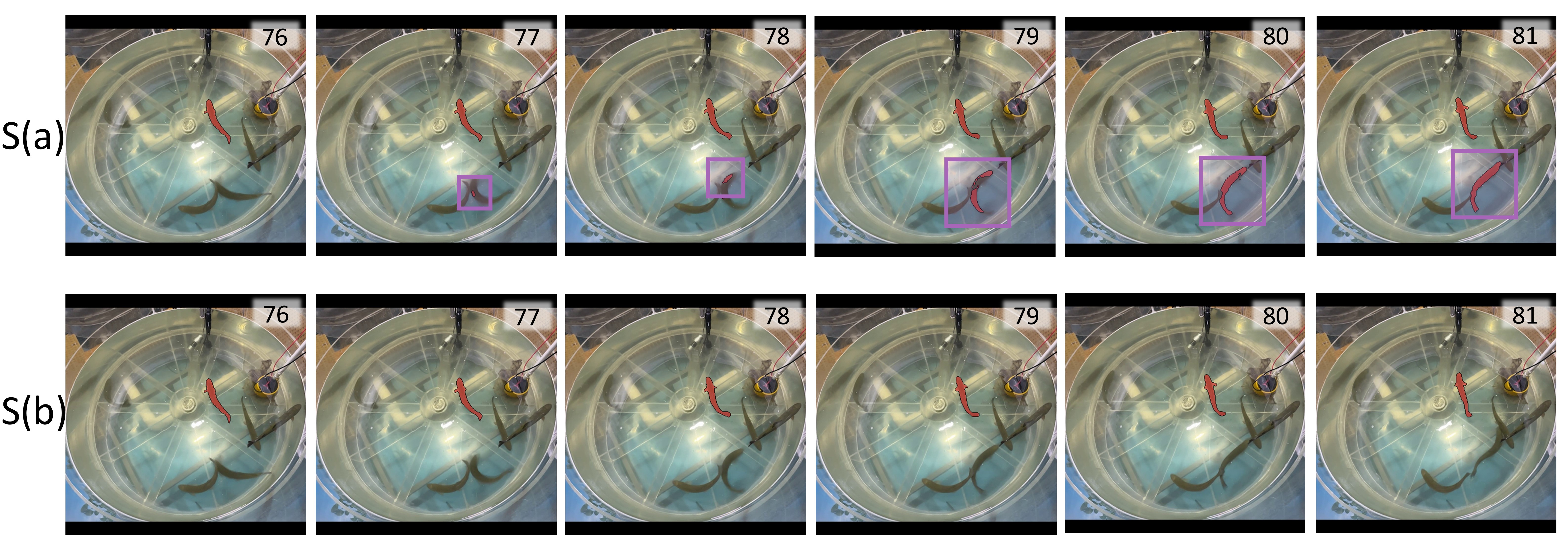}
\caption{
Ablation study of the mask block filter.
S(a) displays the segmentation result of the baseline model, while S(b) shows the result after applying the mask block filter. The purple transparent box indicates the target recognition errors in the baseline model during propagation.
}
\label{fig:lab-s}
\end{figure}

As illustrated in \Cref{fig:lab-s}, section S(a) shows the results of single-object interactive video segmentation in the first propagation round.
It can be observed that the model without the filter exhibited target recognition errors after frame 75 from the interactive frame. As the frame count increased, errors rapidly accumulated, causing the extent of recognition errors to grow.
However, with the addition of the mask block filter, the model results in S(b) retained high recognition accuracy past frame 75, completely preventing recognition errors.
This suggests that in long video sequences, when recognition errors accumulate to trigger error diffusion, the mask block filter indeed plays a crucial role, effectively containing further error spread and enhancing model stability.

\clearpage

As depicted in \Cref{fig:lab-m}, section M(o) shows the results of multi-object video interactive segmentation in the first propagation, where a recognition error occurs at frame 75, followed by another for the second target at frame 114.
Specific details of the recognition errors are shown in M(o)-L. With the mask block filter added, the results and local details are displayed in M(a) and M(a)-L.
These results indicate that the mask block filter is more effective in correcting propagation errors in longer video sequences.

\subsubsection{Noise filter}
\begin{figure}[ht!]
\centering
\includegraphics[
  width=0.7\textwidth]
{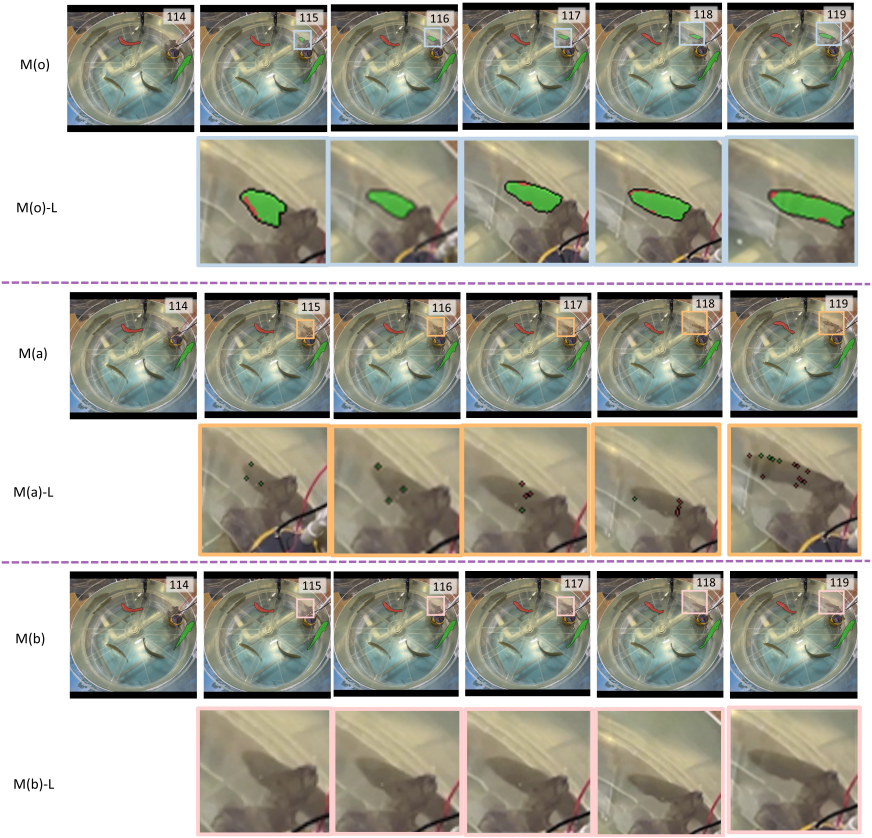}
\caption{
Qualitative ablation study: Mask Block Filter and Noise Filter. 
M(o) and M(o)-L display the segmentation results of the baseline model and an enlarged view of local errors. M(a) and M(a)-L show the interactive results and local magnification after adding the mask block filter. M(b) and M(b)-L illustrate the segmentation results after adding both the mask block filter and noise filter.
}
\label{fig:lab-m}
\end{figure}

As shown in \cref{tab:self-res-s}, in the Fish-DAVIS short video validation, adding the noise filter increased accuracy by 0.604\% over the baseline.
In the Fish-DAVIS long video validation shown in \cref{tab:self-res-l}, the accuracy improvement was even more substantial, with an increase of 17.6\% over the baseline after adding the noise filter.
During noise filtering, the noise filter is essential for removing erroneous masks and reducing noise.
As shown in \Cref{fig:lab-m}, M(a) and M(a)-L display instances of erroneous mask noise particles that may appear after the mask block filter removes incorrect mask blocks, with these particles also displaying an accumulation effect during propagation.
At this stage, the noise filter takes effect, as shown in M(b) and M(b)-L, where the median filter effectively removes these small erroneous particles. After noise filtering, the model's recognition performance improves significantly, preventing further error propagation.

\subsubsection{Rule-Based Thresholding Method}
As shown in \cref{tab:self-res-s}, in the Fish-DAVIS short video dataset, defining the mask filter's distance threshold through the rule-based thresholding method improved accuracy by 0.04\% compared to manually set thresholds based on prior knowledge.
In the Fish-DAVIS long video validation shown in \cref{tab:self-res-l}, defining the mask filter’s distance threshold through the rule-based thresholding method improved accuracy by 9.4\% compared to the manually set threshold.
By setting the optimal distance threshold using the rule-based thresholding method, accuracy reached its highest level.

\begin{figure}[ht!]
\centering
\includegraphics[
  width=0.6\textwidth]
{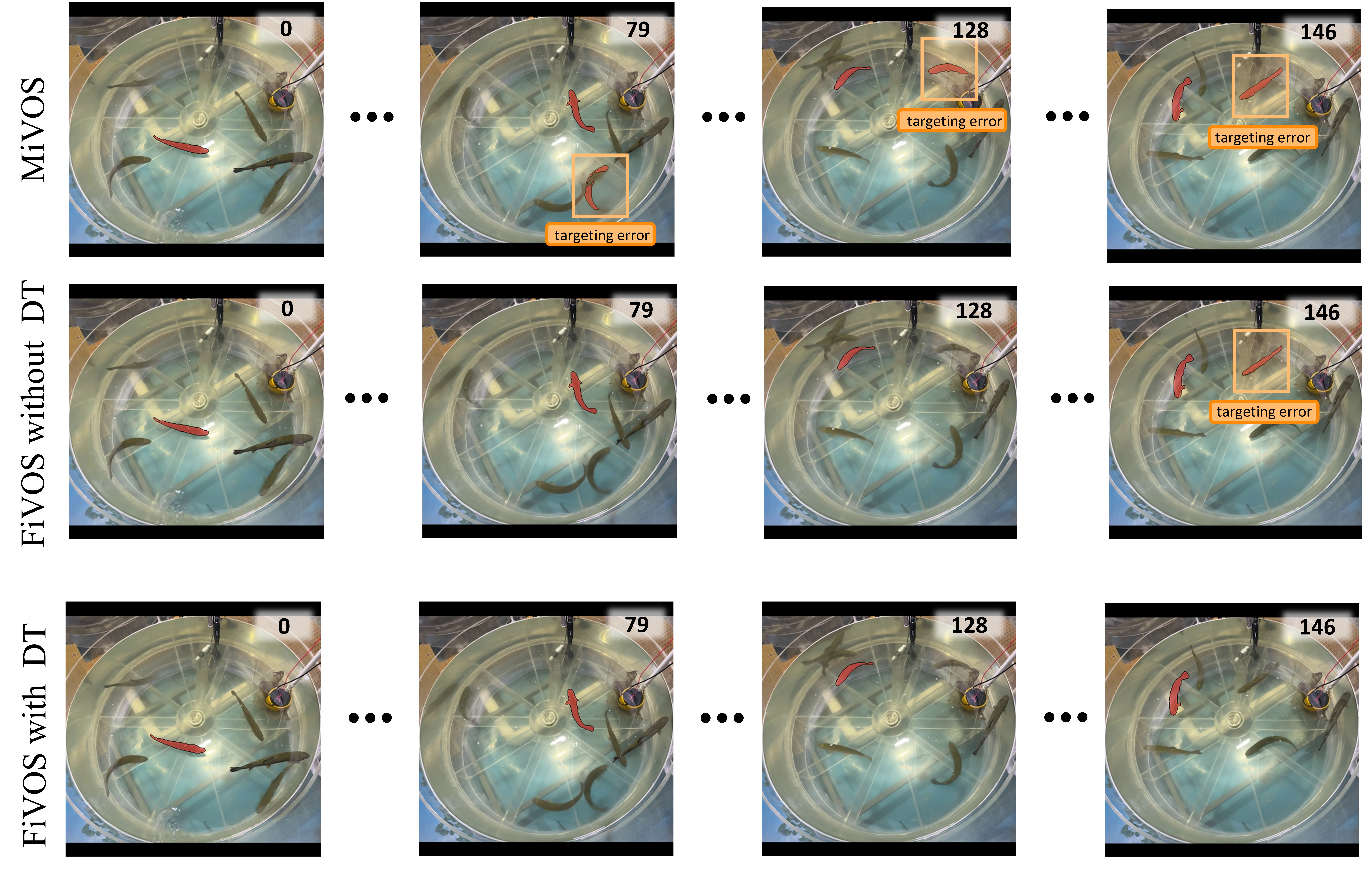}
\caption{
Ablation study of rule-based thresholding. Compared to manually set empirical thresholds (in the second row), the rule-based thresholding method (in the third row) significantly enhances filter performance.}
\label{fig:compare-dt}
\end{figure}

As shown in \Cref{fig:compare-dt}, in the long sequence video interaction results, while the manually set prior distance threshold successfully filtered out erroneous masks at frame 79, recognition errors reappeared after frame 146.
In contrast, the distance threshold derived using the rule-based thresholding method successfully filtered out erroneous masks at both frame 79 and frame 146.
This result demonstrates that, compared to manually set prior thresholds, the rule-based thresholding method offers a more precise distance threshold for identifying erroneous mask blocks, significantly improving the model’s stability and reliability in long-sequence video validation and achieving optimal segmentation performance.

\subsubsection{Generalization Verification}
As shown in \cref{tab:davis-res}, the ablation experiment results on the public dataset DAVIS \cite{perazzi2016benchmark} indicate a slight decrease in accuracy by 0.529\% and 0.792\% after incorporating our filtering scheme.
This slight decrease is mainly due to the high contrast between targets and background in the public dataset and the low similarity between targets, which limits the filter’s effect on this dataset.
Additionally, in certain cases, the noise filter may inadvertently remove a small amount of information, causing a slight effect on the mask results.
However, this impact is minimal, statistically insignificant, and does not affect the overall performance of the method.
Notably, the mask block filter and noise filter were originally designed to address high intra-class similarity targets in complex aquatic environments, where this filtering mechanism shows its advantages.
In practical applications, a slight change in accuracy represents a reasonable trade-off for enhanced robustness.

\begin{table}[!htbp]
\centering
\caption{\label{tab:davis-res}Results of validation on the DAVIS dataset.} 
\label{tab:open data res} 
\begin{tabular}{ccc} 
\toprule 
 & AUC-J \&F & J \&F \\
\midrule 
MiVOS & \textbf{81.450} & \textbf{81.833} \\
MIVOS+mask block filter & 80.921 & 81.392  \\
MiVOS+mask+noise & 80.658 & 81.127 \\
\bottomrule 
\end{tabular}
\end{table}

\section{Discussion}
\subsection{Robustness of the Method}
\begin{figure}[ht!]
\centering
\includegraphics[
  width=\textwidth]
{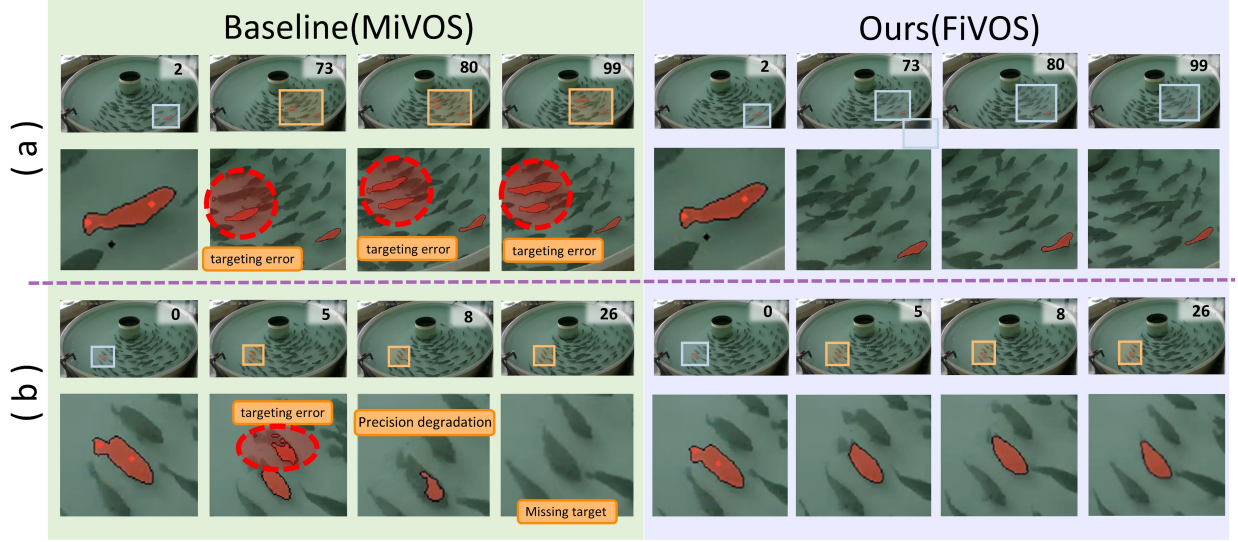}
\caption{Experimental comparison in real farming scenarios}
\label{fig:case1}
\end{figure}

To verify the robustness of the proposed FiVOS method, we conducted additional experiments in more complex scenarios.
First, we collected a video dataset of fish activity in a real aquaculture setting, which includes 150 farmed fish.
We then applied the dataset to both the baseline model (MiVOS) and the proposed FiVOS model, with the experimental results shown in \cref{fig:case1}.
The left section with the green background presents the segmentation results of MiVOS, while the right section with the purple background shows the segmentation results of FiVOS.
We randomly selected two fish as interaction objects, with the results shown in \cref{fig:case1} (a) and (b), respectively.
The first row shows screenshots of the algorithm's segmentation results, and the second row shows the corresponding zoomed-in detail images.
From the detail images, it can be seen that the MiVOS model exhibits significant target confusion and target loss during segmentation; in contrast, the FiVOS model shows stable segmentation results without target confusion.

\begin{figure}[ht!]
\centering
\includegraphics[
  width=\textwidth]
{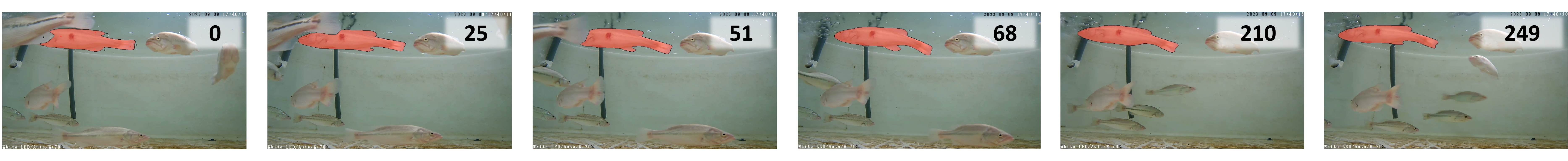}
\caption{Underwater scenario experimental results schematic}
\label{fig:case2}
\end{figure}

To further validate the robustness of our method in underwater scenarios, we collected real underwater fish video data and conducted additional validation experiments. Specifically, we performed interactive segmentation on diseased fish in the videos. The experimental results in \cref{fig:case2} show that our method maintains excellent segmentation performance despite partial occlusion of foreground targets by the background at frames 0, 25, and 51.

These additional experiments effectively demonstrate the validity of our method in real aquaculture environments and underwater scenarios, where it maintains robust performance even under cluttered backgrounds and fish occlusions.

\subsection{Limitations}
\begin{figure}[ht!]
\centering
\includegraphics[
  width=\textwidth]
{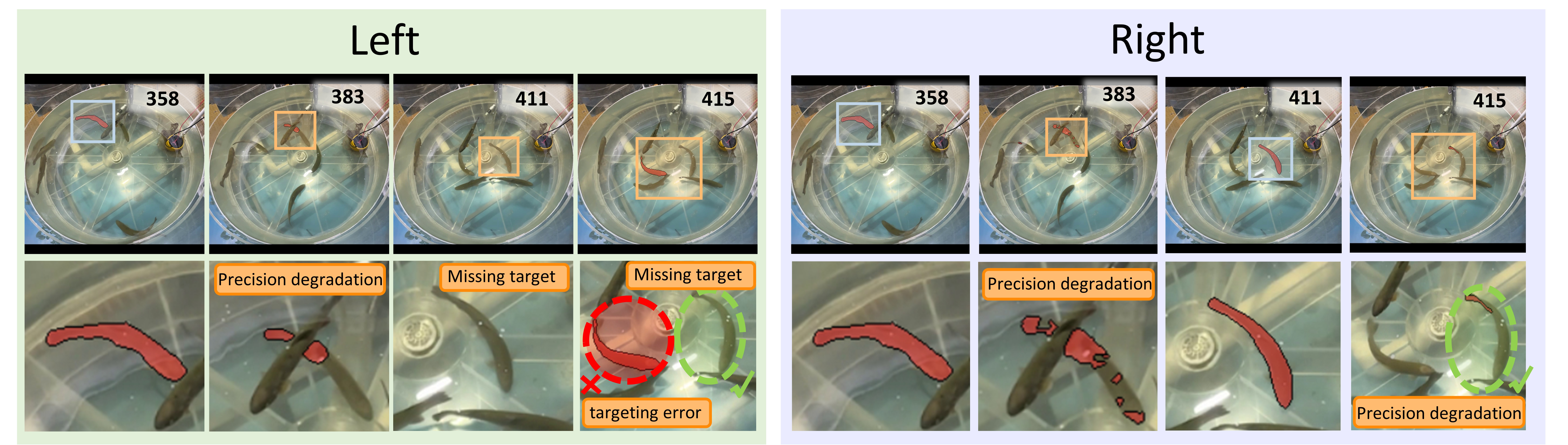}
\caption{
Visualization of FIVOS limitations: suboptimal performance of the fusion module during multiple rounds of interaction.}
\label{fig:error}
\end{figure}

The designed FiVOS has certain limitations, primarily reflected in the following aspects:



As illustrated in \cref{fig:error}, during the first round of interactions, target disappearance caused by occlusion occurs between frames 358 and 411, resulting in misrecognition at frame 415 (as shown in the left part).
Subsequent interaction on frame 411 removes the erroneous mask block in frame 415; however, the recognition of the correct target remains suboptimal (as shown in the right part).
This occurs because, in subsequent interaction propagation, the system increasingly relies on the fusion module, making it difficult for the user to flexibly remove unwanted mask blocks via the filter.
Owing to the characteristics of the fusion module, mask synthesis and correction in later interactions become more complex, limiting the filter's precision and subsequently reducing system accuracy and user control.

\begin{figure}[ht!]
\centering
\includegraphics[
  width=0.7\textwidth]
{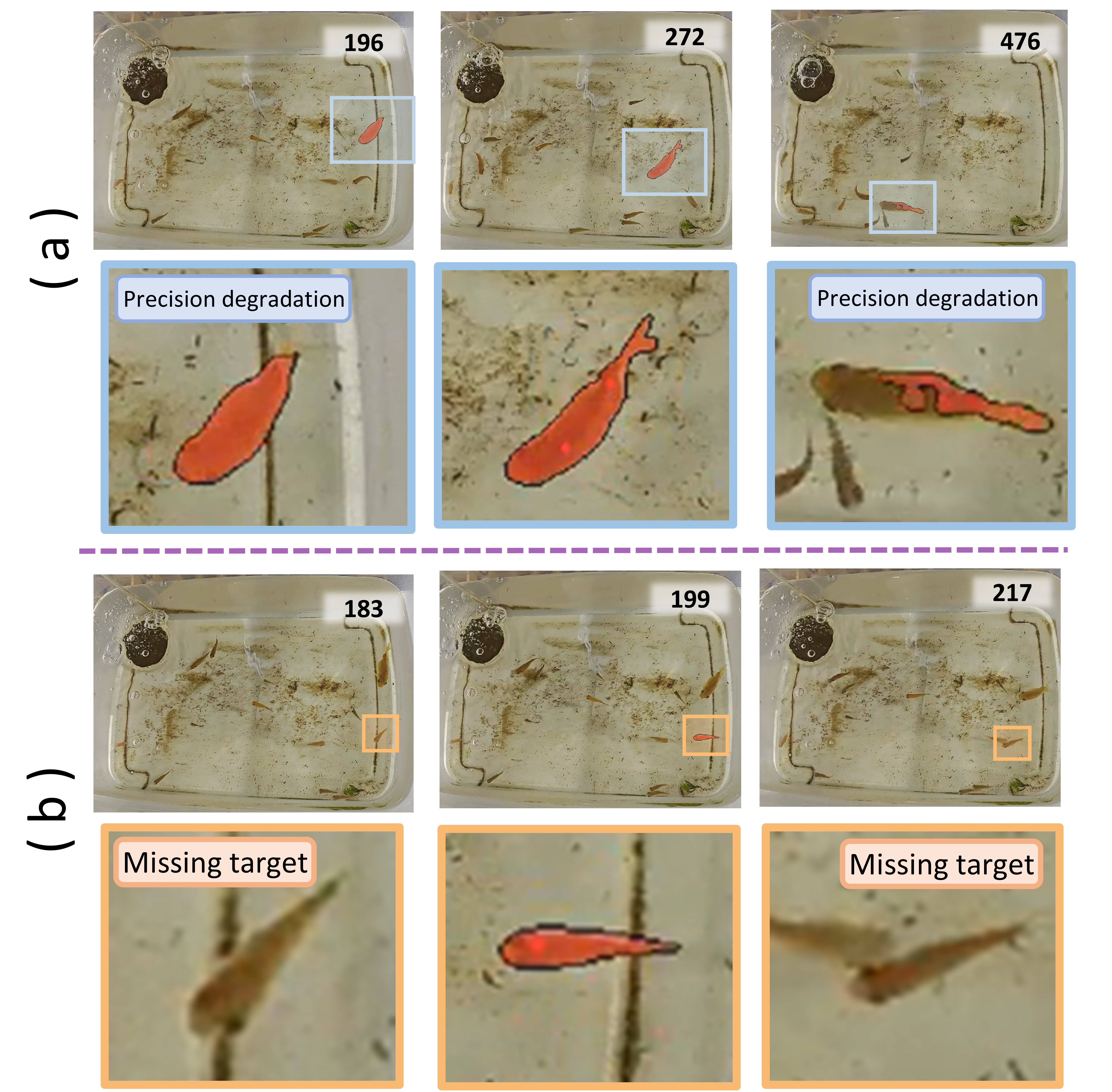}
\caption{Limitations of small target segmentation. (a) Segmentation result when the target occupies 0.8\% of the image. (b) Segmentation result when the target occupies 0.2\% of the image.}
\label{fig:limit-small}
\end{figure}

Our system also exhibits significant limitations when handling small targets. As shown in \cref{fig:limit-small}, we used juvenile goldfish as the validation experiment subject. In \cref{fig:limit-small} (a), the target occupies only 0.8\% of the image. Despite the algorithm’s general satisfactory performance after propagation, some masks still exhibit noticeable detail loss and instability in mask recognition. In \cref{fig:limit-small} (b), the target size is further reduced to 0.2\%, at which point the algorithm almost completely fails due to insufficient spatial resolution for effective feature extraction, leading to complete target loss.

\subsection{Application Prospects}

\begin{figure}[ht!]
\centering
\includegraphics[width=\textwidth]{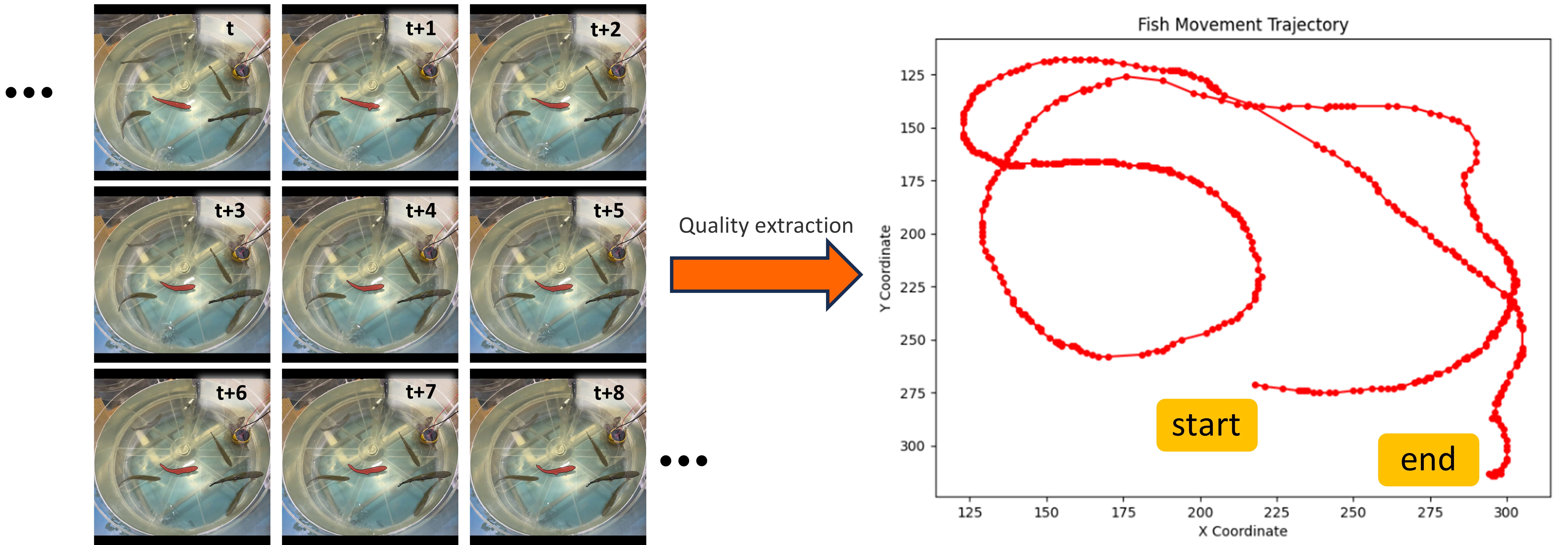}
\caption{Visualization of fish movement trajectory. On the left is a sample of mask frames, where the centroid of each frame’s mask is extracted and sequentially plotted on a coordinate axis, then connected in temporal order.The right side displays the final trajectory plot
.}
\label{fig:lab-apply}
\end{figure}

We used the mask results generated by FiVOS to visualize fish movement trajectories. The specific steps are as follows: first, for each frame's mask, we calculate its centroid position.
The centroid, representing the geometric center of all pixel points within the predicted target mask, effectively indicates the fish’s position in that frame.
By calculating the geometric moments of the mask image, we can quickly compute the centroid coordinates \((x_t, y_t)\) , where \(x_t\) and \(y_t\) are the x and y coordinates of the mask in frame t.
Next, we collect and record the centroid coordinates for each frame. This step ensures a comprehensive understanding of the fish’s movement trajectory across the entire video sequence.
Finally, we connect the collected centroid coordinates in temporal sequence on the coordinate axis, forming a trajectory path map of the fish’s movement in the water, as shown in \Cref{fig:lab-apply}.

This visualization approach allows for an intuitive display of fish swimming paths in water, providing data support for behavioral research and intelligent aquaculture management.
In the trajectory plot shown in \Cref{fig:lab-apply}, adjacent trajectory points are spaced 0.03 seconds apart. Increasing distance between points reflects an increase in swimming speed, potentially related to feeding or stress behaviors; decreasing point distances indicate slowing movement, where centroid distribution narrows or stagnates, possibly indicating poor health (such as hypoxia or death). This dynamic display reveals the fish’s movement state and behavioral characteristics.
Through trajectory data analysis, researchers can further identify normal activities such as foraging and resting, detect abnormal behaviors in a timely manner, and explore regional preferences, providing insights for optimizing feed distribution and water management.
Additionally, long-term tracking of trajectory changes can reveal fish responses to environmental factors (e.g., temperature, water quality). For instance, cold-water species like rainbow trout may cluster in cooler zones when temperature distribution is uneven, suggesting that environmental adjustments might be necessary.
Overall, trajectories derived from segmentation results reveal fish movement patterns and provide precise data support for intelligent aquaculture management.

\section{Conclusions}

To address the difficulty of extracting specific target fish representation information in modern aquaculture due to high intra-class similarity, this paper proposes a novel interactive video object segmentation method, FiVOS.
We designed a simple and effective mask block filter. This filter employs a flexible hard-processing approach, precomputing the maximum centroid distance of mask blocks in an aquaculture dataset and setting it as the filter’s threshold. By calculating the distance between centroids of mask blocks in adjacent frames, it filters out erroneous masks early, resolving misidentification issues due to high intra-class similarity in aquaculture scenarios and addressing the accumulation of recognition errors in the STM model.
Additionally, we observed that after the mask block filter removes certain erroneous mask blocks, there may be a probability of error noise emerging. To address this, we introduced a noise filter after the mask block filter to further reduce the impact of minor noise on the recognition results.
Finally, comparative experiments with current state-of-the-art models were conducted on the fish aquaculture dataset and the DAVIS public dataset, along with ablation studies to further analyze the contributions of the model.
Experimental results demonstrate that our filter not only improves performance on the fish aquaculture dataset but also enhances long-term recognition accuracy and robustness. On the public dataset, its performance is also comparable to that of the original model.

Future work can focus on the following aspects: First, improving the design of the filter to enable better collaboration with the fusion module during subsequent interaction rounds, thereby enhancing the precision and user control in multi-round interactions. 
Secondly, we will expand the diversity of our dataset by incorporating a broader range of fish species and color variations, aiming to enhance the algorithm's effectiveness and user experience across a wider range of application scenarios. Finally, we plan to integrate multi-scale feature fusion and high-resolution input pipelines into our algorithm to enhance segmentation accuracy for small targets, accommodate objects of various sizes.

\section{Acknowledgments}

This paper was supported by The National Natural Science Foundation of China (NO.322
\\
73188), The Key Research and Development Plan of the Ministry of Science and Technology (NO.2022YFD2001700), and Hainan Seed Industry Laboratory (B23H10004).

\bibliographystyle{elsarticle-num} 
\bibliography{references_}

@inproceedings{long2015fully,
  title={Fully convolutional networks for semantic segmentation},
  author={Long, Jonathan and Shelhamer, Evan and Darrell, Trevor},
  booktitle={Proceedings of the IEEE conference on computer vision and pattern recognition},
  pages={3431--3440},
  year={2015}
}

@inproceedings{nam2016learning,
  title={Learning multi-domain convolutional neural networks for visual tracking},
  author={Nam, Hyeonseob and Han, Bohyung},
  booktitle={Proceedings of the IEEE conference on computer vision and pattern recognition},
  pages={4293--4302},
  year={2016}
}

@article{zhang2020automatic,
  title={Automatic fish population counting by machine vision and a hybrid deep neural network model},
  author={Zhang, Song and Yang, Xinting and Wang, Yizhong and Zhao, Zhenxi and Liu, Jintao and Liu, Yang and Sun, Chuanheng and Zhou, Chao},
  journal={Animals},
  volume={10},
  number={2},
  pages={364},
  year={2020},
  publisher={MDPI}
}

@article{zhao2024multilevel,
  title={A multilevel lightweight fish respiratory frequency measurement method--Segmentation instead of detection},
  author={Zhao, Shili and Lu, Jiamin and Zhang, Song and Li, Xuefei and Shi, Chen and Li, Daoliang and Zhao, Ran},
  journal={Aquacultural Engineering},
  volume={107},
  pages={102470},
  year={2024},
  publisher={Elsevier}
}

@article{vabo2021automatic,
  title={Automatic interpretation of salmon scales using deep learning},
  author={Vab{\o}, Rune and Moen, Endre and Smoli{\'n}ski, Szymon and Huseb{\o}, {\AA}se and Handegard, Nils Olav and Malde, Ketil},
  journal={Ecological Informatics},
  volume={63},
  pages={101322},
  year={2021},
  publisher={Elsevier}
}

@article{yang2021computer,
  title={Computer vision models in intelligent aquaculture with emphasis on fish detection and behavior analysis: a review},
  author={Yang, Ling and Liu, Yeqi and Yu, Huihui and Fang, Xiaomin and Song, Lihua and Li, Daoliang and Chen, Yingyi},
  journal={Archives of Computational Methods in Engineering},
  volume={28},
  pages={2785--2816},
  year={2021},
  publisher={Springer}
}

@article{zhao2018modified,
  title={Modified motion influence map and recurrent neural network-based monitoring of the local unusual behaviors for fish school in intensive aquaculture},
  author={Zhao, Jian and Bao, Weijun and Zhang, Fengdeng and Zhu, Songming and Liu, Ying and Lu, Huanda and Shen, Mingwei and Ye, Zhangying},
  journal={Aquaculture},
  volume={493},
  pages={165--175},
  year={2018},
  publisher={Elsevier}
}

@article{vaswani2017attention,
  title={Attention is all you need},
  author={Vaswani, A},
  journal={Advances in Neural Information Processing Systems},
  year={2017}
}

@article{liu2024dp,
  title={DP-FishNet: Dual-path Pyramid Vision Transformer-based underwater fish detection network},
  author={Liu, Yang and An, Dong and Ren, Yinjie and Zhao, Jian and Zhang, Chi and Cheng, Jiahui and Liu, Jincun and Wei, Yaoguang},
  journal={Expert Systems with Applications},
  volume={238},
  pages={122018},
  year={2024},
  publisher={Elsevier}
}

@INPROCEEDINGS{7026059,
  author={Chuang, Meng-Che and Hwang, Jenq-Neng and Kuo, Fang-Fei and Shan, Man-Kwan and Williams, Kresimir},
  booktitle={2014 IEEE International Conference on Image Processing (ICIP)}, 
  title={Recognizing live fish species by hierarchical partial classification based on the exponential benefit}, 
  year={2014},
  volume={},
  number={},
  pages={5232-5236},
  doi={10.1109/ICIP.2014.7026059}}

@article{wang2022real,
  title={Real-time detection and tracking of fish abnormal behavior based on improved YOLOV5 and SiamRPN++},
  author={Wang, He and Zhang, Song and Zhao, Shili and Wang, Qi and Li, Daoliang and Zhao, Ran},
  journal={Computers and Electronics in Agriculture},
  volume={192},
  pages={106512},
  year={2022},
  publisher={Elsevier}
}

@article{zhao2022lightweight,
  title={A lightweight dead fish detection method based on deformable convolution and YOLOV4},
  author={Zhao, Shili and Zhang, Song and Lu, Jiamin and Wang, He and Feng, Yu and Shi, Chen and Li, Daoliang and Zhao, Ran},
  journal={Computers and Electronics in Agriculture},
  volume={198},
  pages={107098},
  year={2022},
  publisher={Elsevier}
}

@article{li2024yolo,
  title={YOLO-FD: An accurate fish disease detection method based on multi-task learning},
  author={Li, Xuefei and Zhao, Shili and Chen, Chunlin and Cui, Hongwu and Li, Daoliang and Zhao, Ran},
  journal={Expert Systems with Applications},
  volume={258},
  pages={125085},
  year={2024},
  publisher={Elsevier}
}

@article{sun2022water,
  title={Water quality monitoring method based on G-RepVGG and fish movement behavior},
  author={Sun, Longqing and Wang, Boning and Wang, Jiayu and others},
  journal={Transactions of the Chinese Society for Agricultural Machinery},
  volume={53},
  number={sup 2},
  pages={210--218},
  year={2022}
}

@misc{vieira2023assessment,
  title={Assessment of age and growth in fishes},
  author={Vieira, Ana Rita},
  journal={Fishes},
  volume={8},
  number={10},
  pages={479},
  year={2023},
  publisher={MDPI}
}

@article{mandal2024role,
  title={Role of artificial intelligence (AI) in fish growth and health status monitoring: A review on sustainable aquaculture},
  author={Mandal, Arghya and Ghosh, Apurba Ratan},
  journal={Aquaculture International},
  volume={32},
  number={3},
  pages={2791--2820},
  year={2024},
  publisher={Springer}
}

@article{papadakis2012computer,
  title={A computer-vision system and methodology for the analysis of fish behavior},
  author={Papadakis, Vassilis M and Papadakis, Ioannis E and Lamprianidou, Fani and Glaropoulos, Alexios and Kentouri, Maroudio},
  journal={Aquacultural engineering},
  volume={46},
  pages={53--59},
  year={2012},
  publisher={Elsevier}
}

@article{delcourt2013video,
  title={Video multitracking of fish behaviour: a synthesis and future perspectives},
  author={Delcourt, Johann and Deno{\"e}l, Mathieu and Ylieff, Marc and Poncin, Pascal},
  journal={Fish and Fisheries},
  volume={14},
  number={2},
  pages={186--204},
  year={2013},
  publisher={Wiley Online Library}
}

@article{zhao2024vision,
  title={Vision-based dual network using spatial-temporal geometric features for effective resolution of fish behavior recognition with fish overlap},
  author={Zhao, Haixiang and Wu, Yuankai and Qu, Keming and Cui, Zhengguo and Zhu, Jianxin and Li, Hao and Cui, Hongwu},
  journal={Aquacultural Engineering},
  volume={105},
  pages={102409},
  year={2024},
  publisher={Elsevier}
}

@inproceedings{pedersen20203d,
  title={3d-zef: A 3d zebrafish tracking benchmark dataset},
  author={Pedersen, Malte and Haurum, Joakim Bruslund and Bengtson, Stefan Hein and Moeslund, Thomas B},
  booktitle={Proceedings of the IEEE/CVF conference on computer vision and pattern recognition},
  pages={2426--2436},
  year={2020}
}

@inproceedings{sharif2015simple,
  title={A simple approach to count and track underwater fishes from videos},
  author={Sharif, Md Haidar and Galip, Feyza and Guler, Adil and Uyaver, Sahin},
  booktitle={2015 18th international conference on computer and information technology (ICCIT)},
  pages={347--352},
  year={2015},
  organization={Ieee}
}

@article{ye2016behavioral,
  title={Behavioral characteristics and statistics-based imaging techniques in the assessment and optimization of tilapia feeding in a recirculating aquaculture system},
  author={Ye, Zhangying and Zhao, Jian and Han, Zhiying and Zhu, Songming and Li, Jianping and Lu, Huanda and Ruan, Yunjie},
  journal={Transactions of the ASABE},
  volume={59},
  number={1},
  pages={345--355},
  year={2016},
  publisher={American Society of Agricultural and Biological Engineers}
}

@article{li2024tfmft,
  title={TFMFT: Transformer-based multiple fish tracking},
  author={Li, Weiran and Liu, Yeqiang and Wang, Wenxu and Li, Zhenbo and Yue, Jun},
  journal={Computers and Electronics in Agriculture},
  volume={217},
  pages={108600},
  year={2024},
  publisher={Elsevier}
}

@inproceedings{redmon2016you,
  title={You only look once: Unified, real-time object detection},
  author={Redmon, J},
  booktitle={Proceedings of the IEEE conference on computer vision and pattern recognition},
  year={2016}
}

@article{benard2017interactive,
  title={Interactive video object segmentation in the wild},
  author={Benard, Arnaud and Gygli, Michael},
  journal={arXiv preprint arXiv:1801.00269},
  year={2017}
}

@inproceedings{xu2016deep,
  title={Deep interactive object selection},
  author={Xu, Ning and Price, Brian and Cohen, Scott and Yang, Jimei and Huang, Thomas S},
  booktitle={Proceedings of the IEEE conference on computer vision and pattern recognition},
  pages={373--381},
  year={2016}
}

@inproceedings{caelles2017one,
  title={One-shot video object segmentation},
  author={Caelles, Sergi and Maninis, Kevis-Kokitsi and Pont-Tuset, Jordi and Leal-Taix{\'e}, Laura and Cremers, Daniel and Van Gool, Luc},
  booktitle={Proceedings of the IEEE conference on computer vision and pattern recognition},
  pages={221--230},
  year={2017}
}

@inproceedings{oh2019fast,
  title={Fast user-guided video object segmentation by interaction-and-propagation networks},
  author={Oh, Seoung Wug and Lee, Joon-Young and Xu, Ning and Kim, Seon Joo},
  booktitle={Proceedings of the IEEE/CVF Conference on Computer Vision and Pattern Recognition},
  pages={5247--5256},
  year={2019}
}

@inproceedings{heo2020interactive,
  title={Interactive video object segmentation using global and local transfer modules},
  author={Heo, Yuk and Jun Koh, Yeong and Kim, Chang-Su},
  booktitle={Computer Vision--ECCV 2020: 16th European Conference, Glasgow, UK, August 23--28, 2020, Proceedings, Part XVII 16},
  pages={297--313},
  year={2020},
  organization={Springer}
}

@inproceedings{miao2020memory,
  title={Memory aggregation networks for efficient interactive video object segmentation},
  author={Miao, Jiaxu and Wei, Yunchao and Yang, Yi},
  booktitle={Proceedings of the IEEE/CVF Conference on Computer Vision and Pattern Recognition},
  pages={10366--10375},
  year={2020}
}

@inproceedings{sofiiuk2020f,
  title={f-brs: Rethinking backpropagating refinement for interactive segmentation},
  author={Sofiiuk, Konstantin and Petrov, Ilia and Barinova, Olga and Konushin, Anton},
  booktitle={Proceedings of the IEEE/CVF Conference on Computer Vision and Pattern Recognition},
  pages={8623--8632},
  year={2020}
}

@inproceedings{lu2020video,
  title={Video object segmentation with episodic graph memory networks},
  author={Lu, Xiankai and Wang, Wenguan and Danelljan, Martin and Zhou, Tianfei and Shen, Jianbing and Van Gool, Luc},
  booktitle={Computer Vision--ECCV 2020: 16th European Conference, Glasgow, UK, August 23--28, 2020, Proceedings, Part III 16},
  pages={661--679},
  year={2020},
  organization={Springer}
}

@inproceedings{oh2019video,
  title={Video object segmentation using space-time memory networks},
  author={Oh, Seoung Wug and Lee, Joon-Young and Xu, Ning and Kim, Seon Joo},
  booktitle={Proceedings of the IEEE/CVF international conference on computer vision},
  pages={9226--9235},
  year={2019}
}

@inproceedings{luiten2018premvos,
  title={Premvos: Proposal-generation, refinement and merging for video object segmentation},
  author={Luiten, Jonathon and Voigtlaender, Paul and Leibe, Bastian},
  booktitle={Asian conference on computer vision},
  pages={565--580},
  year={2018},
  organization={Springer}
}

@inproceedings{cheng2021modular,
  title={Modular interactive video object segmentation: Interaction-to-mask, propagation and difference-aware fusion},
  author={Cheng, Ho Kei and Tai, Yu-Wing and Tang, Chi-Keung},
  booktitle={Proceedings of the IEEE/CVF Conference on Computer Vision and Pattern Recognition},
  pages={5559--5568},
  year={2021}
}

@inproceedings{perazzi2016benchmark,
  title={A benchmark dataset and evaluation methodology for video object segmentation},
  author={Perazzi, Federico and Pont-Tuset, Jordi and McWilliams, Brian and Van Gool, Luc and Gross, Markus and Sorkine-Hornung, Alexander},
  booktitle={Proceedings of the IEEE conference on computer vision and pattern recognition},
  pages={724--732},
  year={2016}
}

@article{cheng2021rethinking,
  title={Rethinking space-time networks with improved memory coverage for efficient video object segmentation},
  author={Cheng, Ho Kei and Tai, Yu-Wing and Tang, Chi-Keung},
  journal={Advances in Neural Information Processing Systems},
  volume={34},
  pages={11781--11794},
  year={2021}
}

@article{fiordelmondo2020improvement,
  title={Improvement of the water quality in rainbow trout farming by means of the feeding type and management over 10 years (2009--2019)},
  author={Fiordelmondo, Elisa and Magi, Gian Enrico and Mariotti, Francesca and Bakiu, Rigers and Roncarati, Alessandra},
  journal={Animals},
  volume={10},
  number={9},
  pages={1541},
  year={2020},
  publisher={MDPI}
}

@inproceedings{cheng2022xmem,
  title={{XMem}: Long-Term Video Object Segmentation with an Atkinson-Shiffrin Memory Model},
  author={Cheng, Ho Kei and Alexander G. Schwing},
  booktitle={ECCV},
  year={2022}
}

@InProceedings{oh2018fast,
author = {Oh, Seoung Wug and Lee, Joon-Young and Sunkavalli, Kalyan and Kim, Seon Joo},
title = {Fast Video Object Segmentation by Reference-Guided Mask Propagation},
booktitle = {The IEEE Conference on Computer Vision and Pattern Recognition (CVPR)},
year = {2018}
}

\end{document}